\documentclass[10pt,journal,compsoc]{IEEEtran}

\ifCLASSOPTIONcompsoc
  \usepackage[nocompress]{cite}
\else
  \usepackage{cite}
\fi

\ifCLASSINFOpdf
\else
\fi

\usepackage{stfloats}

\usepackage[table,xcdraw]{xcolor}
\usepackage{cite}
\usepackage{amsmath,amssymb,amsfonts}
\usepackage{algorithmic}
\usepackage{graphicx}
\usepackage{textcomp}
\usepackage{tabularx}
\usepackage{makecell}
\usepackage{pgfplots}
\usepackage{multirow}
\usepackage{amssymb}
\usepackage{pifont}
\newcommand{\cmark}{\ding{51}}
\newcommand{\xmark}{\ding{55}}
\usepackage{soul}
\usepackage{subcaption}

\usepackage{changepage}
\usepackage{pdflscape}
\usepackage{comment} 
\usepackage{multicol}
\usepackage{float}
\usepackage{breqn}
\usepackage{arydshln}

\makeatletter
\newcommand*{\addFileDependency}[1]{%
  \typeout{(#1)}
  \@addtofilelist{#1}
  \IfFileExists{#1}{}{\typeout{No file #1.}}
}
\makeatother

\usepackage{hyperref}

\begin{document}
\title{StratDef: Strategic Defense Against Adversarial Attacks in ML-based Malware Detection}

\author{Aqib~Rashid, Jose~Such%
\IEEEcompsocitemizethanks{\IEEEcompsocthanksitem The authors are with the Department of Informatics, King's College London, Strand, London WC2R 2LS, United Kingdom.\protect\\
E-mail: \{aqib.rashid, jose.such\}@kcl.ac.uk}%
}

\IEEEtitleabstractindextext{%

\begin{abstract}
Over the years, most research towards defenses against adversarial attacks on machine learning models has been in the image recognition domain. The ML-based malware detection domain has received less attention despite its importance. Moreover, most work exploring these defenses has focused on several methods but with no strategy when applying them. In this paper, we introduce StratDef, which is a strategic defense system based on a moving target defense approach. We overcome challenges related to the systematic construction, selection, and strategic use of models to maximize adversarial robustness. StratDef dynamically and strategically chooses the best models to increase the uncertainty for the attacker while minimizing critical aspects in the adversarial ML domain, like attack transferability. We provide the first comprehensive evaluation of defenses against adversarial attacks on machine learning for malware detection, where our threat model explores different levels of threat, attacker knowledge, capabilities, and attack intensities. We show that StratDef performs better than other defenses even when facing the peak adversarial threat. We also show that, of the existing defenses, only a few adversarially-trained models provide substantially better protection than just using vanilla models but are still outperformed by StratDef.
\end{abstract}

}

\maketitle

\IEEEdisplaynontitleabstractindextext

\IEEEpeerreviewmaketitle

\IEEEraisesectionheading{\section{Introduction}\label{sec:introduction}}

The advantages of ML models in fields such as image recognition, anomaly detection, and malware detection are undisputed, as they can offer unparalleled performance on large, complex datasets \cite{he2015delving,chio2018machine}. Nevertheless, such models are vulnerable to adversarial examples \cite{Eykholt_2018_CVPR, evtimov2017robust} which are inputs that are intentionally designed to induce a misclassification. Resilience against adversarial examples is essential and critical, with much work having been carried out in the image recognition domain to defend against adversarial examples \cite{chakraborty2018adversarial, tramer2017ensemble, xu2017feature, wang2016random, xie2017mitigating, sengupta2018mtdeep}. However, these defenses are often less effective in the more constrained malware detection domain \cite{chen2018automated, pierazzi2020problemspace}. Worryingly, out of the papers published in the last seven years on adversarial machine learning, approximately only 50 out of 3,000+ relate to the malware detection domain \cite{carlinilist}. In fact, a recent survey that took an initial step towards evaluating defenses applied to this domain painted a bleak picture \cite{pods}.

While complete security is difficult to achieve, a system's goal should be to control the attack surface as much as possible to thwart attacks. Existing defenses in this regard are based on a variety of techniques \cite{chakraborty2018adversarial} such as adversarial training \cite{szegedy2013intriguing, tramer2017ensemble}, gradient-based approaches \cite{papernot2016distillation, tramer2017ensemble}, feature-based approaches \cite{xu2017feature, wang2016random} and randomization-based approaches \cite{xie2017mitigating, sengupta2018mtdeep} with mixed success. Despite these multiple research efforts at developing defenses, there is little work approaching the problem from a strategic perspective. For this purpose, in other areas of cybersecurity, a moving target defense (MTD) is employed that continually varies itself to increase the uncertainty and complexity for the attacker, making reconnaissance and targeted attacks less successful \cite{mtddef, cho2020toward}. There are numerous ways that an MTD can vary itself, with some approaches having been applied to adversarial ML before \cite{sengupta2018mtdeep, qian2020ei, amich2021morphence, song2019moving, 9833895, additional1, additional2}, but not in the malware detection domain nor in the depth we explore. Namely, we provide a method for constructing a strategic defense that embraces the key areas of model construction, model selection, and optimizer selection for a strategic MTD.

In this paper, we present our defense method, StratDef. We investigate how a \emph{strategized defense} can offer better protection against adversarial attacks in the malware detection domain. We suggest methods to combat attacks strategically based on an MTD approach (rather than relying on a single model) by considering various factors that have not been explored in detail before, such as model heterogeneity, threat level, and information available about the attacker. Furthermore, we investigate various dimensions of a strategic MTD, such as what, how, and when it should adapt itself given the current environment it is operating within. %
Our goal is to make the job of the attacker more difficult by increasing the uncertainty and complexity of the problem. Moreover, existing defenses do not consider a \emph{systematic} model selection process for the ensemble \cite{sengupta2018mtdeep,qian2020ei,wang2020mtdnnf,shahzad2013comparative,yerima2018droidfusion, amich2021morphence, song2019moving, 9833895, additional1, additional2}. This process is non-trivial and must deal with selecting the constituent models of the ensemble and then how to strategically use them. We demonstrate promising approaches for model selection and the subsequent, strategic use of the selected models for offering reliable predictions and protection against adversarial ML attacks. We further provide an experimental evaluation across Android and Windows to demonstrate the fragility of individual models and defenses compared with StratDef.

The main contributions of our work can be summarized as follows:
\begin{itemize}
    \item We propose the first strategic defense against adversarial attacks in the malware detection domain. Our defense, StratDef, is based on an MTD approach where we propose different strategic and heuristically-driven methods for determining what, how, and when a defense system should move to achieve a high degree of adversarial robustness. This includes key steps related to model selection and the development of strategies.
    
    \item We offer a detailed evaluation of existing defensive approaches to demonstrate the necessity of a strategized approach by comparing existing defenses with ours. That is, we consider the constraints and characteristics of this domain in a proper manner, unlike prior evaluations. The results show that our strategized defense can increase accuracy by 50+\% in the most adverse conditions in both Android and Windows malware.
    
    \item We are the first to evaluate how a strategized defense based on MTDs fares against a variety of attackers, such as gray-box attackers with limited knowledge, black-box attackers with zero-knowledge, and attackers who only use adversarial examples generated with Universal Adversarial Perturbations (UAPs).
    
\end{itemize}

The rest of this paper is organized as follows. Section~\ref{sec:background} provides the background and puts StratDef in the context of related work. In Section~\ref{sec:applicationandthreatmodel}, we define the threat model used in our work. In Section~\ref{sec:methodandprocedure}, we provide details about our defensive method, StratDef. In Sections \ref{sec:experimentalsetup} and \ref{sec:evaluation}, we present our experimental setting and results, respectively. We present a discussion on our findings in Section~\ref{sec:discussion} and we conclude in Section~\ref{sec:conclusion}.

\section{Background \& related work}
\label{sec:background}
\noindent{\textbf{Adversarial ML and Malware.}} Machine learning is increasingly being relied on for the detection of malware. An ML-based malware detection classifier must be accurate and robust, as well as precise with good recall. The quality of such a classifier hinges on the features used during the training procedure \cite{grosse2016, al2018adversarial, rosenberg2018generic, demontis2017yes}. %
For software, the process of feature extraction is used to parse a software executable into its feature representation. Accordingly, the use of APIs, libraries, system calls, resources, or the accessing of network addresses, as well as the actual code are parsed into discrete, binary feature vectors to represent the presence or absence of a feature. Then, together with the class labels (i.e., benign and malware), models such as neural networks are trained on the feature vectors to classify unseen inputs.

However, the problem with using ML-based detection models is that they are vulnerable to \emph{adversarial examples} \cite{szegedy2013intriguing}. These are inputs to ML models that are intentionally designed to fool a model by having the model output the attacker's \emph{desired prediction} through an \emph{evasion attack} \cite{papernot2018sok}. For example, an image of a panda may be incorrectly classified as a gibbon \cite{goodfellow2014explaining} or a truly malicious executable may be misclassified as benign \cite{grosse2017adversarial}. In some cases, an adversarial example generated for a particular model may also evade another model too \cite{szegedy2013intriguing} due to \emph{transferability}. 

To generate a new adversarial example for an image, an evasion attack can be performed by using one of several attacks from prior work, which perturb values in the feature vector representing the image (i.e., its pixels) \cite{goodfellow2014explaining, al2018adversarial, yang2017malware, kurakin2016adversarial, carlini2017towards, chen2020hopskipjumpattack, madry2017towards, papernot2016limitations}. However, these attacks cannot be applied directly to the malware detection domain as they make perturbations to continuous feature vectors without due consideration for the domain's constraints. When generating an adversarial example for the malware detection domain, the malicious functionality must be preserved (in the feature-space) and the feature vector must remain discrete \cite{10.1145/3473039, rosenberg2020query, demetrio2019explaining, grosse2017statistical, pierazzi2020problemspace, ebrahimi2020binary, sewak2021adversarialuscator, song2020mab}. For example, a feature representing an API call (e.g., $GetTempPath()$) cannot be perturbed continuously (e.g., $GetTempPath() + 0.001$). Instead, an entirely new feature must be used \cite{rosenberg2020query, pierazzi2020problemspace} that offers the same functionality. This increases the complexity of working in this domain. To deal with this, when perturbations are applied by an attack, it must be ensured that they are \emph{permitted} and \emph{proper} to cater to the constraints imposed by this domain. For this, we present a method to achieve a lower bound of functionality-preservation in the feature-space (see Section~\ref{sec:experimentalsetup} later). Recent work by Demetrio et al. \cite{10.1145/3473039} has suggested that malware samples constructed as adversarial examples may not always be functional. To this end, the RAMEN framework has been proposed, which can alter malware structure without affecting the functionality. This approach has shown promise in adversarial attacks.

In Table~\ref{table:malwarerelatedlit}, we provide an overview of the related literature on the subject of adversarial ML and malware. In particular, the literature can be separated by the domain, the aim of the paper (e.g., proposing a new defense or attack), and whether the work is conducted in the feature-space (where ML-based algorithms operate using feature vectors) or the problem-space (where real-world objects such as software executables exist).

\begin{table}[!htbp]
\centering
\scalebox{0.85}{
\begin{tabular}{p{0.3\linewidth}|p{0.7\linewidth}}
\hline
Category & Literature \\
\hline
Attacks & \cite{rosenberg2020query, rosenberg2018generic, 10.1145/3473039, demetrio2019explaining, demetrio2021functionality, pierazzi2020problemspace, ebrahimi2020binary, song2020mab, grosse2016, grosse2017adversarial, labaca2021universal, demontis2017yes, al2018adversarial, hu2017generating, suciu2019exploring, demetrio2022practical, chen2019android, anderson2018learning} \\
\hdashline[0.5pt/5pt]
Defenses & \cite{grosse2017statistical, li2020enhancing, yerima2018droidfusion, chen2020training2, maiorca2019towards} \\
\hdashline[0.5pt/5pt]
Surveys/Evaluations & \cite{pods, grosse2016, grosse2017adversarial, 10.1145/3484491, maiorca2019towards, ling2023adversarial, aryal2021survey, li2021arms} \\
\hdashline[0.5pt/5pt]
Feature-space & \cite{rosenberg2020query, rosenberg2018generic, 10.1145/3473039, demetrio2019explaining, demetrio2021functionality, ebrahimi2020binary, song2020mab, pods, grosse2016, grosse2017adversarial, grosse2017statistical, labaca2021universal, demontis2017yes, al2018adversarial, hu2017generating, li2020enhancing, yerima2018droidfusion, chen2020training2, 10.1145/3484491, suciu2019exploring, ling2023adversarial, chen2019android, anderson2018learning} \\
\hdashline[0.5pt/5pt]
Problem-space & \cite{10.1145/3473039, demetrio2021functionality, pierazzi2020problemspace, labaca2021universal, demetrio2022practical} \\
\hdashline[0.5pt/5pt]
Android domain & \cite{pierazzi2020problemspace, pods, grosse2016, grosse2017adversarial, grosse2017statistical, labaca2021universal, demontis2017yes, li2020enhancing, yerima2018droidfusion, aryal2021survey, li2021arms, chen2019android} \\
\hdashline[0.5pt/5pt]
Windows domain & \cite{rosenberg2020query, rosenberg2018generic, 10.1145/3473039, demetrio2019explaining, demetrio2021functionality, pierazzi2020problemspace, ebrahimi2020binary, song2020mab, labaca2021universal, al2018adversarial, suciu2019exploring, demetrio2022practical, ling2023adversarial, aryal2021survey, li2021arms, anderson2018learning} \\
\hdashline[0.5pt/5pt]
PDF domain & \cite{chen2020training2, maiorca2019towards, aryal2021survey, li2021arms} \\
\hline
\end{tabular}
}
\caption{Overview of related literature on adversarial ML in malware.}
\label{table:malwarerelatedlit}
\end{table}

\noindent{\textbf{Defenses.}} To deal with the threat of adversarial ML, several defenses have been proposed, mainly for the image recognition domain, with mixed success \cite{pods, carlini2019evaluating}. These include a range of techniques such as adversarial training \cite{szegedy2013intriguing, tramer2017ensemble}, gradient-based approaches \cite{papernot2016distillation, tramer2017ensemble}, feature-based approaches \cite{xu2017feature, wang2016random} and randomization-based approaches \cite{xie2017mitigating}. For example, Papernot et al. proposed defensive distillation \cite{papernot2016distillation} which involves utilizing different-sized neural networks to improve the generalization of the main model, though Stokes et al. \cite{stokes2017attack} found this to be ineffective when applied to the malware detection domain. Wang et al. proposed random feature nullification, which decreases the attacker's chances of using features that are important to the model \cite{wang2016random}. This is only effective if the attacker chooses to perturb features randomly as well \cite{pods}. Xie et al. \cite{xie2017mitigating} also proposed a randomization-based defense, though this has been shown to be ineffective by Athayle et al. \cite{athalye2018obfuscated}. Another approach is to mask and obfuscate gradients, though this has been found ineffective in later work\cite{papernot2017practical}. Podschwadt et al. found that adversarial training (first proposed in \cite{szegedy2013intriguing}) is a potentially effective defense method \cite{pods} though it cannot protect against unknown and unseen threats. However, a limitation of their work is that they do not sufficiently consider the constraints of this domain. Moreover, in our work, we identify and validate some issues with this method. For example, adversarial training introduces additional complexities such as determining which model to choose as the base model, what to train on, and how much to train. However, as we show, our StratDef approach assists with this, as it helps to select the most promising models and then choose between them strategically at prediction-time. This produces better results than a single model trained adversarially, as shown in Section~\ref{sec:evaluation}.

\noindent{\textbf{Moving Target Defenses.}} In a moving target defense (MTD), the configuration of the defense changes regularly. The key design principles of an MTD include the ``what to move'', the ``how to move'' and the ``when to move'' \cite{cho2020toward}. In the context of adversarial ML, this typically involves \emph{moving} between the ML models used to make predictions. Thus, MTDs can be considered to belong to the family of ensemble defenses. The objective is to make it more challenging for attackers to perform meaningful reconnaissance and successful attacks \cite{cho2020toward}, which will be rendered difficult as the target model will not be static. Different MTD approaches have offered some success in other domains %
\cite{sengupta2018mtdeep, qian2020ei, amich2021morphence, song2019moving}. 
However, MTDs have never been applied to the malware detection domain before. %
To the best of our knowledge, we are the first to explore how an MTD approach can defend against adversarial attacks in the ML-based malware detection domain with our defense, StratDef. StratDef advances the state of the art by embracing the key principles of an MTD. Rather than plainly utilizing an MTD approach with a group of models, StratDef provides an entire framework for generating models, selecting those models systematically, and producing reliable strategies to use those models to offer accurate predictions against legitimate inputs while defending against adversarial examples.

Existing MTD approaches from other domains do not consider various key factors that we explore, such as the challenges related to the systematic construction, selection, and strategic use of models to maximize adversarial robustness. For example, prior MTD-based work only uses small ensembles of models, consisting mainly of DNNs as the constituent models, and varies these DNNs only in their training procedure. We explore how and to what degree the model selection procedure should be heuristically-driven to promote key aspects such as heterogeneity, redundancy, and to minimize the effect of transferability of adversarial examples across models.  Moreover, unlike other defenses, StratDef can give consideration to information available about its operating environment to provide an adapted and tailored response based on the current threat level. 

Next, we introduce the threat model used in our work, followed by a detailed description of our defense, StratDef.

\section{Threat Model}
\label{sec:applicationandthreatmodel}

Feature-based ML malware detection is a domain that has been widely explored in previous work \cite{severi2021explanation, yang2017malware, suciu2019exploring, grosse2016, grosse2017adversarial, biggio2013evasion, apruzzese2023realgradients}. Our work focuses on the same well-established threat model concerning the evasion of such malware detection classifiers. Examples of such systems are antivirus with an ML component, or a Machine-Learning-as-a-Service (MLaaS) platform that offers this functionality (e.g., VirusTotal).

\begin{figure}[!ht]
\centering
    \scalebox{0.65}{
        \includegraphics{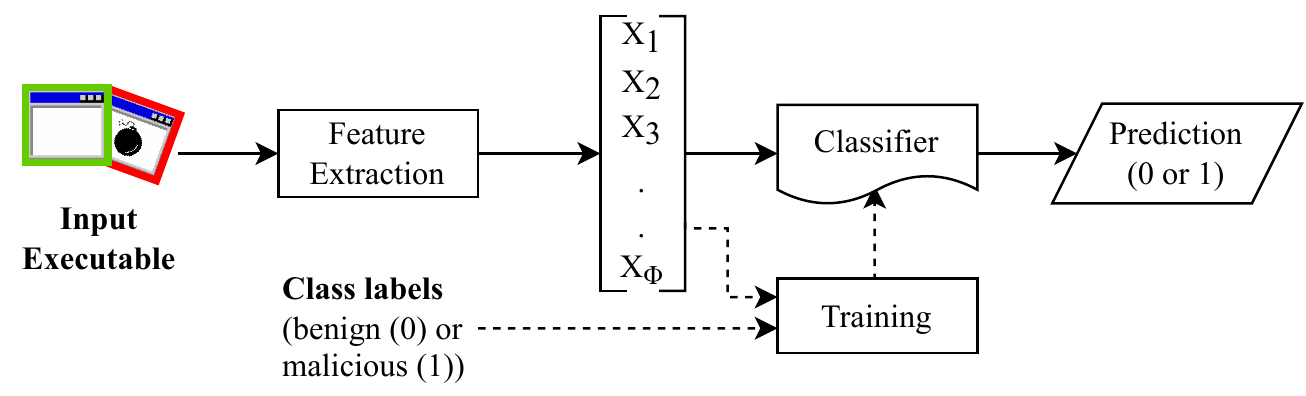}
    }
    \caption{Overview of a malware detection classifier: the dataset provides features that are processed into feature vectors of size \(\Phi\). Class labels are assigned to each input sample. Feature vectors and class labels form training data.}
  \label{figure:malwareclassifierschema}
\end{figure}

\noindent\textbf{Application Model.} The target model of the attack is a binary classification model that predicts whether an input sample belongs to the benign or malware class. To construct such a classifier for malware detection, executables are represented as binary feature vectors. For this purpose, datasets often provide a comprehensive set of extracted features from real-world executables. With these datasets and features \(1 \ldots \Phi\), we can construct a vector \(X\) for each input sample such that \(X \in \{{0,1}\}^\Phi \). \(X_i = 1\) indicates the presence of feature \(i\) and \(X_i = 0\) indicates its absence. We use the feature vectors and associated class labels to construct binary classification models for malware detection as shown in Figure~\ref{figure:malwareclassifierschema}.

\noindent{\textbf{Attacker's Goal.}}
The attacker's goal is to generate adversarial examples to evade a malware detection classifier. Suppose we have a classifier \(F\), where \(F\colon X \in \{{0,1}\}^\Phi \) and a function \(chk()\) to check the functionality of an input sample. Then, this goal can be summarized as:

\begin{equation}
    chk(X) = chk(X'); F(X) = 1; F(X') = 0
\end{equation}

We use 0 to represent the benign class and 1 for the malware class. The attacker wants to generate an adversarial example \(X'\) that is functionally-equivalent to the original malware sample \(X\), but is predicted as \emph{benign} by \(F\). %

\noindent{\textbf{Attacker Knowledge \& Capabilities.}}
For the majority of the paper, we model all attackers who interact with StratDef in a gray-box setting with \emph{limited knowledge} about the target model, like previous work \cite{papernot2018sok, laskov2014practical, santana2021detecting, biggio2018wild}. This may represent a scenario such as an insider threat or one where model data has been leaked. In our threat model, attackers have access to the same training data as the target model and knowledge of the feature representation. However, attackers have no knowledge of the parameters, configurations, or constituent models of StratDef nor any other evaluated defenses. Therefore, they must train substitute models using the training data and attack them, in the expectation that the generated adversarial examples will transfer to the target model \cite{szegedy2013intriguing, laskov2014practical, papernot2017practical}. This is based on the well-established idea that adversarial examples for different models can be used to evade the target model \cite{szegedy2013intriguing}. %

Furthermore, we use different scenarios involving attacker capabilities and attack intensities with the goal of studying and evaluating the performance of StratDef under different \emph{threat levels}, like in prior work \cite{carlini2017adversarial, carlini2019evaluating, biggio2018wild, 217486, severi2021explanation, pierazzi2020problemspace}. Attackers may differ in their behavior, the strength and intensity of their attacks, their ability to generate adversarial examples, and more. For deployment, in the absence of any information about the operating environment, StratDef assumes the highest threat level, consisting of the most adverse environment and the strongest attacker. However, if there is information about the operating environment and/or the attackers within it (e.g., through cyber-threat intelligence \cite{shu2018threat,zhu2018chainsmith} or situational awareness), StratDef can use it to provide a more targeted defensive approach. %
Therefore, in our evaluation (see Section~\ref{sec:evaluation} later), we show how StratDef performs against different attacker scenarios and intensities to show the whole range of its capabilities. Nonetheless, for the comparison with other defenses later, we focus on the strongest attacker, as this is the default scenario when no information is available about the attacker or environment.

Additionally, we evaluate StratDef's performance against a black-box attacker with \emph{zero knowledge}, as featured in previous work \cite{ilyas2018black, papernot2018sok, papernot2017practical, rosenberg2018generic, papernot2016transferability, brendel2017decision, chen2020stateful, demetrio2021functionality, ebrahimi2020binary, Li_2020_CVPR}. This attacker only has access to the predictions of StratDef and no other knowledge. For instance, an attacker may not have direct access to a remote model, but may monitor how it responds to different queries. This scenario is especially pertinent nowadays due to the ubiquity of MLaaS platforms. The black-box attacker therefore performs a transferability attack, in which they construct a substitute model as an estimation of the target model by querying it systematically. The substitute model is attacked in the hope that any generated adversarial examples transfer to the target model \cite{szegedy2013intriguing}.

\section{StratDef}
\label{sec:methodandprocedure}
\begin{figure}[!h]
\centering
    \scalebox{0.75}{
        \includegraphics{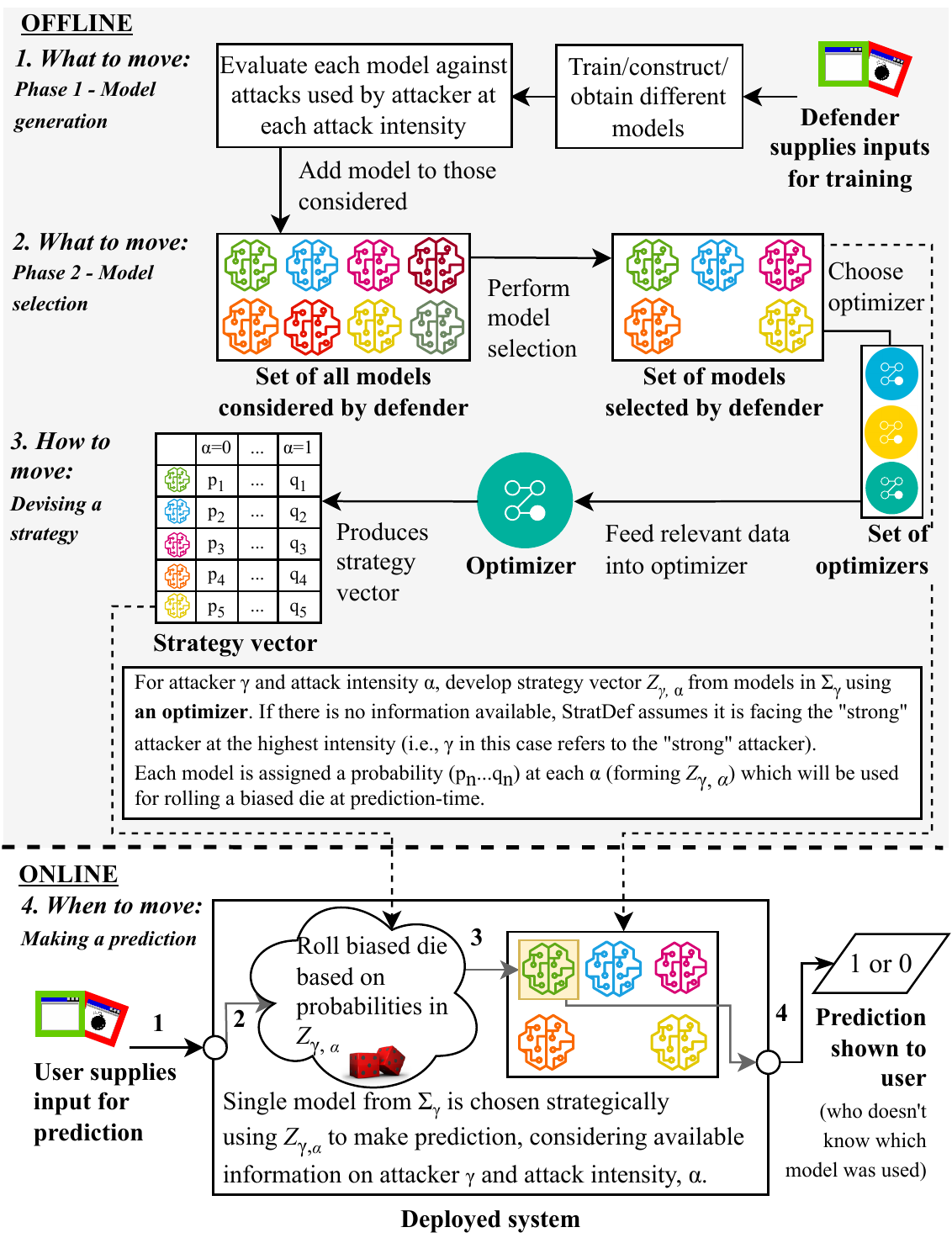}
   }
    \caption{Overview of StratDef.}
  \label{figure:ourmethod}
\end{figure}

In this section, we firstly describe our strategic method, StratDef, at a high level and then provide details about each of its steps. StratDef embraces the three key design principles of an MTD: what to move, how to move, and when to move \cite{cho2020toward, mtddef, wang2020mtdnnf}. We provide a systematic and heuristic model selection system to determine \emph{what to move} considering the current user type, the threat level, and performance metrics.  With this method, StratDef's behavior can be optimized according to the situation; for example, if a particular metric needs to be prioritized, the model selection can be adjusted accordingly (as we describe later). As presented in the overview of StratDef in Figure~\ref{figure:ourmethod}, once models have been selected, we can strategize \emph{how} they will be used. In cybersecurity, an MTD typically cycles through configurations during its deployment. Since StratDef makes predictions for legitimate and adversarial inputs, we use a strategy to choose a model at prediction-time, thereby strategically cycling through the models \emph{when} it is time to move. We explore multiple methods to determine this strategy, ranging from uniform random probabilities, a game-theoretic approach based on Bayesian Stackelberg Games, and a strategic heuristically-driven approach. 

Overall, StratDef provides an entire method for choosing the models to use in the ensemble and then how to use them at prediction-time to serve predictions to users. We next provide details for the key steps of StratDef, which are shown in Figure~\ref{figure:ourmethod}.

\subsection{What to Move: Phase 1 --- Model Generation}
\label{sec:whattomovedescparta}
The first step is conducted offline (that is, at the development stage --- see Figure~\ref{figure:ourmethod}). This consists of generating the models that are going to be considered when forming the StratDef ensemble. Let \(U\) refer to the set of candidate models considered by the defender for inclusion within the ensemble. These models can be constructed by the defender or taken from other work. \(U\) can contain vanilla models as well as models that already incorporate individual defenses (single-model defenses, e.g., adversarial training \cite{szegedy2013intriguing}, random feature nullification \cite{wang2016random}, SecSVM \cite{demontis2017yes}, defensive distillation \cite{papernot2016distillation}). In addition, \(U\) can contain models of different families, such as decision trees, neural networks, random forests, or support vector machines, as well as models of the same family but with different parameters (e.g., neural networks trained adversarially but to different levels). %

\subsection{What to Move: Phase 2 --- Model Selection}
\label{sec:whattomovedescpartb}
Continuing offline, the model selection procedure is performed to produce a strong, heterogeneous ensemble of models to be used. %
Recall that StratDef provides a tailored defense by considering the information it may have about its operating environment. That is, the model selection (and the strategy derived for using those selected models later) is tailored to the operating environment based on any available information about it. However, if no specific information is available, an environment with the highest threat level is assumed (i.e., the strongest attacker and highest attack intensity) with the model selection reflecting this. A benefit of having a tailored model selection (provided that information is available about the environment) is that the StratDef ensemble will be appropriate for the level of threat, potentially reducing the defender's costs (e.g., resources, time required for training, storage). For example, the costs of training highly-robust models could be avoided if StratDef is deployed in a less hostile environment where less robust models might suffice. 

To the best of our knowledge, with our model selection approach, we are the first to offer a flexible method to select models systematically by considering model performance and threat levels. To achieve this model selection system, we \emph{simulate} threat levels by generating adversarial examples and pooling them (we provide details in Sections~\ref{sec:crafting} \& \ref{sec:attackerprofiles} later on how the adversarial examples are generated and pooled).
Each candidate model \(F \in U\) is then evaluated under each threat level using several machine learning metrics. %
This allows us to aggregate the metrics into a \emph{consideration score}, %
which encapsulates the performance of each candidate model at different threat levels.

A higher consideration score for a model indicates its better performance, which may increase its chances of being selected for the StratDef ensemble. The actual formula for the consideration score can vary based on the deployment needs and requirements. For example, a defender may be more interested in minimizing false positives over other metrics (see Section~\ref{sec:trainingmodelsanddefsexp} later for the specific formula we use in our experimental evaluation). Therefore, in Equation~\ref{equation:considerationscoregeneral}, we provide a \emph{general} formula for the consideration score:

\begin{equation}
 \begin{aligned}
    S_{F, \gamma, \alpha} & = \oplus (m^1_{F, \gamma, \alpha}, m^2_{F, \gamma, \alpha}, \ldots, m^n_{F, \gamma, \alpha})
  \end{aligned}
  \label{equation:considerationscoregeneral}
\end{equation}

\(S_{F, \gamma, \alpha}\) refers to the consideration score of a candidate model \(F \in U\) at attack intensity \(\alpha\) against attacker \(\gamma\). That is, a particular combination of the metrics chosen by the defender is considered (e.g., whether metrics are weighted, maximized, minimized, etc.).  For the \(n\) considered metrics, \(m_{F, \gamma, \alpha}\) refers to the metric \(m\) for the candidate model \(F\) at attack intensity \(\alpha\) against attacker \(\gamma\). The defender can choose the metrics and attribute weights to each in whatever manner they require as part of the consideration score. Depending on the situation, one may adjust the considered metrics or use different metrics altogether to produce a new model selection. 

Once the consideration scores are produced, the candidate models are sorted in descending order by their consideration scores at each threat level (i.e., each attack intensity \(\alpha\) and each attacker \(\gamma\)). Essentially, this procedure ensures that the candidate models can be sorted by their performance considering several ML metrics at different threat levels. This drives the model selection method for selecting which candidate models will be part of the StratDef ensemble. %
We explore two different model selection methods that use the consideration scores and other characteristics of the candidate models (and evaluate these methods later in Section~\ref{sec:evaluation}):

\begin{itemize}
    \item \textbf{\textit{Best} selection method} --- This method selects the best-performing models with the aim of maximizing performance across the considered metrics. For each attacker and attack intensity (i.e., each threat level), we select the \(k\) highest-scoring models out of all potential candidates in \(U\). %
    \(k\) is a hyperparameter and can be chosen by the defender according to their requirements and resources (e.g., a defender with greater resources may want to use more models). The models selected \emph{may} be from the same model families, yet they demonstrate some robustness and are therefore chosen to deal with attacks. %

    \item \textbf{\textit{Variety} selection method} --- This method aims to reduce transferability among the selected models by enforcing diversity in the model selection. The highest-scoring model from each model family in \(U\) is selected at each attack intensity against each attacker. The number of models selected per attack intensity is equal to the number of model families. 
\end{itemize}

Once the selections are made according to the method, the model selections for each attack intensity \(\alpha\) for each attacker \(\gamma\) are pooled together and represented by the set \(\Sigma_{\gamma}\), so that it contains all the models selected across all attack intensities for the attacker \(\gamma\). Recall that if no information is available about the operating environment, \(\gamma\) represents a strong attacker, with the most capable models selected to deal with this kind of environment. This model selection process offers a systematic yet flexible approach to the defender, allowing them to adjust the considered metrics, which would adapt StratDef to their needs. For example, the defender could modify the formulation of consideration score, the candidate models to be considered, or the model selection method itself.%

\subsection{How to Move: Devising a Strategy}
\label{sec:howtomovedesc}

So far, StratDef has selected an ensemble of models for each threat level (i.e., each attacker and attack intensity) that are considered most suitable according to various criteria. The models selected across the attack intensities for each attacker (\(\gamma\)) have then been pooled together into the set \(\Sigma_{\gamma}\). The next step is to determine \emph{how} these models will actually be used at prediction-time to serve predictions to users. For this, an optimizer is used to strategize how each model in \(\Sigma_{\gamma}\) (the models that were selected in the previous step) will be chosen at prediction-time by StratDef for making predictions. %
This step takes place offline and corresponds to choosing an optimizer for developing the movement strategy, as in Figure~\ref{figure:ourmethod}.

Each optimizer produces a global strategy vector \(Z_{\gamma}\) for an attacker \(\gamma\) using data about the models. The probability of choosing each model from \(\Sigma_{\gamma}\) at attack intensity \(\alpha\) against attacker \(\gamma\) at prediction-time is contained within \(Z_{\gamma, \alpha}\). This means that the strategy chooses from the most suitable models to make the prediction by adapting to the attack intensity and the attacker type (i.e., the threat level). With the probabilities in \(Z_{\gamma}\), a biased die is thrown at prediction-time to select a model for making the prediction. %

In order to devise the movement strategies and the strategy vector \(Z_{\gamma}\), we explore three optimizers and evaluate them later in Section~\ref{sec:evaluation}. We next provide a description of how a strategy vector is produced by each optimizer that we consider in this work.

\noindent{\textbf{Game-Theoretic Optimizer (GT).}} 
We can model the problem of adversarial examples as a min-max game between an attacker and a defender, following the well-established concept of \emph{Security Games}. This has been successfully applied to various areas of both physical security and cybersecurity \cite{paruchuri2008playing}. Specifically, in our problem setting, the attacker is trying to maximize the loss of the classifier, while the defender is trying to minimize it.

Hence, we model the interaction between the defender (\(\mathcal{D}\)) and the user --- who can be either a legitimate user (\(\mathcal{L}\)) or an attacker (\(\gamma\)) --- as a Bayesian Stackelberg Game \cite{tambe2011security}. The defender is the leader, and the user is the follower. The defender aims to maximize its expected reward over its switching strategy and the strategy vector played by the followers. We produce payoff matrices for each game between the defender and each user. 

The game between the defender and the attacker is modelled as a constant-sum game like previous work in other domains \cite{sengupta2018mtdeep, qian2020ei}. In prior related work, the utilities in game-theoretic formulations of adversarial ML have been based on the evasion rates of each attack (the attacker's possible move) against each model (the defender's possible move), but without any consideration for how many adversarial examples are produced. In the malware detection domain, attacks may be more or less effective because of domain-specific constraints. Additionally, a stronger attacker can generate a greater number of more evasive adversarial examples than a weaker attacker. Therefore, we instead use a \emph{normalized evasion rate} as the utility value in the game-theoretic formulation in order to encapsulate information about the number of adversarial examples generated as well as the evasion rate. We provide details of the procedure below.

Let \(\Omega_{\tau, S}\) represent the set of adversarial examples generated by an attack \(\tau\) for some substitute model \(S\) that is used to generate adversarial examples (see Sections~\ref{sec:crafting} \& \ref{sec:attackerprofiles} later for implementation and experimental instances):

\begin{enumerate}
    \item Evaluate each set of adversarial examples \(\Omega_{\tau, S}\) against each selected model \(F \in \Sigma_{\gamma}\) to obtain the evasion rate. That is, measure the proportion of adversarial examples from \(\Omega_{\tau, S}\) that evade the model \(F\).
   
    \item Compute the normalized evasion rate (\(R_{\tau, S, F}\)) to reflect the evasive capability of the set \(\Omega_{\tau, S}\) against model \(F\). For this, multiply the number of adversarial examples in the set by the evasion rate and normalize between 0 and 100. A constant-sum game (\(= 100\)) is the frequent setup for the game between an attacker and defender \cite{sengupta2018mtdeep, qian2020ei}.
      
    \item Produce payoff matrices, where the defender is the row player, for each game by calculating rewards:
    \begin{enumerate}
        \item For the constant-sum game between \(\mathcal{D}\) and \(\gamma\), the attacker's reward is equal to the normalized evasion rate \(R_{\tau, S, F}\). The defender's reward, because it is a constant-sum game, is therefore equal to \(100 - R_{\tau, S, F}\). %
        \item For the game between \(\mathcal{D}\) and \(\mathcal{L}\), the reward for both players is equal to the accuracy of the model \(F\) on a clean test set (i.e., the defender's possible move).
    \end{enumerate}
    
    \item Feed both payoff matrices into a Bayesian Stackelberg solver (such as \cite{paruchuri2008playing, gurobi}) along with the attack intensities.
    This produces a strategy vector \(Z_{\gamma, \alpha}\) containing the probability of playing (i.e., selecting) each model \(F \in \Sigma_{\gamma}\) against attacker \(\gamma\) at attack intensity \(\alpha\) as \(p({F, \gamma, \alpha})\).
\end{enumerate}

In the optimization problem, \(\alpha\) is a hyperparameter modelled as a trade-off between accuracy on legitimate and adversarial inputs corresponding to the attack intensity. The optimization problem may result in a pure strategy (where only a single model is chosen for predictions) or a mixed strategy (where there is a choice between multiple models). A pure strategy can be produced when one of the models is more robust than others. %
At \(\alpha=0\), StratDef is only concerned with accuracy on legitimate inputs, and therefore a pure strategy of the most accurate model is produced.

\noindent{\textbf{Strategic Ranked Optimizer (Ranked).}} 
With this optimizer approach, we use the consideration scores for each model in the set \(\Sigma_{\gamma, \alpha}\) (i.e., the models selected for attacker \(\gamma\) at attack intensity \(\alpha\)) to produce a strategy vector. At \(\alpha=0\), a pure strategy consisting of the most accurate model is produced. Meanwhile, for \(\alpha > 0\), each model in the set is sorted by its consideration score. A rank is then assigned to each model in the sorted set, with the lowest-scoring model having a rank of 1. The rank increases as the model scores increase. Based on this, each model is assigned a probability in \(Z_{\gamma, \alpha}\) as per Equation~\ref{equation:strategiceqn}:

\begin{equation}
 \begin{aligned}
    p({F, \gamma, \alpha}) & = \frac{r_{F, \gamma, \alpha}}{\sum_{G \in \Sigma_{\gamma, \alpha}} r_{G, \gamma, \alpha}} \\
  \end{aligned}
  \label{equation:strategiceqn}
\end{equation}

\(r_{F, \gamma, \alpha}\) is the rank of model \(F\) at attack intensity \(\alpha\) against attacker \(\gamma\). In other words, the probability of a model \(F\) being selected is its rank divided by the sum of all ranks. Therefore, we assign the highest probability of being selected to the highest-scoring model. In \(Z_{\gamma, \alpha}\), a probability of 0 is assigned to models that are not in \(\Sigma_{\gamma, \alpha}\). In other words, if a model was not selected at a particular attack intensity, it will have a probability of 0 in the strategy vector. This approach will always generate a mixed strategy at every attack intensity except \(\alpha=0\).

\noindent{\textbf{Uniform Random Strategy (URS).}} 
This approach assigns a uniform probability to each model in $\Sigma_{\gamma}$ and only acts as a baseline for comparing with the other approaches, as it is not expected to give the best performance. The uniform random strategy approach maximizes the uncertainty for the attacker with regard to the model that is selected at prediction-time. Thus, the probability is calculated according to Equation~\ref{equation:urseqn}:

\begin{equation}
 \begin{aligned}
    p({F, \gamma, \alpha}) & = \frac{1}{|\Sigma_{\gamma}|} \\
  \end{aligned}
  \label{equation:urseqn}
\end{equation}

\subsection{When to Move: Making a Prediction}
\label{sec:whentomovedesc}
After the offline generation and selection of the best models as well as the creation of the strategies to move between the selected models --- that is, the strategy vector \(Z_{\gamma}\) to move between models in the set \(\Sigma_{\gamma}\) --- StratDef is now ready to be deployed online and start making predictions. %
As per Figure~\ref{figure:ourmethod}, when a user requests a prediction, StratDef will choose a model from \(\Sigma_{\gamma}\) to pass the input to by rolling a biased die using the probabilities in the strategy vector \(Z_{\gamma, \alpha}\) in real-time. The chosen model will actually make the prediction that will be returned to the user. In the absence of information about the threat level of the environment, StratDef will assume it is facing the strong attacker at the highest attack intensity. Because the actual model that is used by StratDef to make each prediction will be chosen dynamically, the user will have it difficult to know how predictions are being produced, let alone which model was used at prediction-time. Therefore, our hypothesis is that if the previous steps are performed systematically following our method, StratDef will offer sound and robust predictions, while revealing minimal information about itself.%

Next, we show how StratDef performs better than existing defenses in the malware detection domain in the face of adversarial examples. In the following section, we provide details of the experimental setup we consider for the evaluation, together with details about how we generate adversarial examples.

\section{Experimental Setup}
\label{sec:experimentalsetup}

\subsection{Datasets}
\label{sec:experimentalsetupdatasets}
In malware detection, the number of publicly-available, up-to-date datasets is a well-known, general problem, which limits the remits and conclusions of academic work in this domain~\cite{arp2022and, 8949524}. We therefore perform our evaluation with two well-known datasets that cover different application platforms, collection dates and have been widely used in prior work. 

The Android DREBIN dataset \cite{arp2014drebin} consists of 123,453 benign samples and 5,560 malware samples collected between 2010 and 2012. There is a total of eight feature families consisting of extracted static features ranging from permissions, API calls, hardware requests and URL requests. DREBIN contains several malware families such as FakeInstaller, DroidKungFu and Plankton; we refer the reader to \cite{arp2014drebin} for a more detailed discussion. To keep our dataset balanced, we use 5,560 samples from each class (benign and malware), resulting in a total of 11,120 samples with 58,975 unique features. Meanwhile, the Windows SLEIPNIR dataset \cite{al2018adversarial} consists of 19,696 benign samples and 34,994 malware samples collected prior to 2018. As our work is in the feature-space, SLEIPNIR is used to represent Windows out of simplicity because it offers a convenient binary feature-space, enabling a clearer comparison between the Android and Windows datasets. The features of this dataset are derived from API calls in PE files parsed into a vector representation by the LIEF library \cite{LIEF, al2018adversarial}. We use 19,696 samples from each class, resulting in a total of 39,392 samples with 22,761 unique features. Similar to recent publications \cite{demontis2017yes, grosse2017adversarial}, and for completeness, we use the maximum features for each dataset. Both datasets are transformed into a binary feature-space with each input sample transformed into a feature vector representation. The datasets are initially split using an 80:20 ratio for training and test data using the Pareto principle. After this, the training data is further split using an 80:20 ratio to produce training data and validation data. This effectively produces a 64:16:20 split, which is a technique that has been widely used before \cite{ma2021partner, ravi2016optimization, li2019few, yao2019hierarchically, du2021metakernel, ye2021train}. 

We consider the established guidelines for performing malware-related research \cite{6234405}. For example, as the models in our evaluation decide whether an input sample is benign or malicious, it is crucial to retain benign samples in the datasets, and we do not need to strictly balance datasets over malware families. Rather, we balance datasets between the positive and negative classes (i.e., benign and malware) and select unique samples from each class to appear in the training and test sets randomly (without any chance of repetition) \cite{pods, al2018adversarial, rosenberg2018generic}.

\subsection{Training Models \& Defenses}
\label{sec:trainingmodelsanddefsexp}
\label{sec:trainingmodels}
\noindent{\textbf{Other Models \& Defenses.}} To construct all models, we use the scikit-learn \cite{scikit-learn}, Keras \cite{chollet2015keras} and Tensorflow \cite{tensorflow2015-whitepaper} libraries. We construct four vanilla models (see Appendix~\ref{appendix:architecutres} for architectures). Vanilla models are the base models for defenses such as ensemble adversarial training \cite{szegedy2013intriguing, tramer2017ensemble}, defensive distillation \cite{papernot2016distillation}, SecSVM \cite{demontis2017yes}, and random feature nullification \cite{grosse2016}. For adversarial training, we train the vanilla models with different sized batches of adversarial examples (ranging from 0.1\% to 25\%) from those generated previously. For example, suppose the size of the test set is 2224 (which is equally split between benign and malware samples); then for a 0.05 model variant (e.g., NN-AT-0.05), we select 56 adversarial examples (i.e., 5\% of half the test set size) and add these to the training set. We then train the vanilla and SecSVM models to produce adversarially-trained models. We found in preliminary work that adversarially training beyond 25\% increases time and storage costs as well as overfitting. We apply defensive distillation to the vanilla NN model, %
while random feature nullification is applied to all vanilla models. The vanilla SVM model acts as the base model for SecSVM. We also compare StratDef with the voting defense. Voting has been applied to other domains \cite{wang2020mtdnnf} and to the malware detection domain \cite{shahzad2013comparative, yerima2018droidfusion}. This is similar to a Multi-Variant Execution Environment (MVEE) where an input sample is fed into multiple models in order to assess divergence and majority voting is used for the prediction \cite{jackson2010multi, jackson2010effectiveness}. We use the same constituent models for the voting defense as for StratDef (and thus the naming conventions are similar). We consider two voting approaches that have been tested in prior work \cite{shahzad2013comparative, yerima2018droidfusion}: \emph{majority voting} and \emph{veto voting}. The better of the two approaches is compared with StratDef. In preliminary work, we discover that veto voting causes higher false positive rate (FPR) in both datasets --- as high as 25\% in DREBIN (see Appendix~\ref{appendix:votingfpr}). This poor performance may be because the voting system is forced to accept any false positive prediction from its constituent models. Therefore, we focus on comparing StratDef with majority voting using the same model selections.

\noindent{\textbf{StratDef.}}
To construct different StratDef configurations, we follow the offline steps described in Section~\ref{sec:methodandprocedure} to construct models and devise strategies. The candidate models are the individual models and defenses trained as described above (except voting). In the StratDef configuration that we use in this work, we aim to maximize the performance and robustness on input samples while minimizing false predictions. Therefore, to achieve this, we use the formula in Equation~\ref{equation:considerationscoreexpv2} for the consideration scores, where we maximize accuracy (ACC), AUC, F1 and minimize FPR and false negative rate (FNR) across the threat levels.

\begin{dmath}
     S_{F, \gamma, \alpha} = ACC_{F, \gamma, \alpha} + F1_{F, \gamma, \alpha} + AUC_{F, \gamma, \alpha} - FPR_{F, \gamma, \alpha} - FNR_{F, \gamma, \alpha}
  \label{equation:considerationscoreexpv2}
\end{dmath}

Recall that this process relates to ``What to Move --- Phase 2'' (see Section~\ref{sec:whattomovedescpartb}). In particular, \(S_{F, \gamma, \alpha}\) is the consideration score of the candidate model \(F\) at attack intensity \(\alpha\) against attacker \(\gamma\). By maximizing the accuracy, F1 and AUC, while minimizing the false positives and negatives, the selected models will offer sound performance, with the use of Equation~\ref{equation:considerationscoreexpv2}. The value of each metric for the candidate model \(F\) at attack intensity \(\alpha\) against attacker \(\gamma\) is represented accordingly. We use all combinations of the Best (with \(k=5\)) and Variety model selection methods with the three optimizers described in Section~\ref{sec:howtomovedesc} to produce six StratDef configurations (see Appendix~\ref{appendix:mtdstrategies} for example strategies developed by StratDef).

\subsection{Practical Considerations \& Characteristics of Adversarial Examples}
\label{sec:considerationsandcharsofcrafting}
When generating adversarial examples in the ML-based malware detection domain, it is vital to ensure that feature vectors remain discrete and that malicious functionality is preserved by limiting the set of allowed perturbations that can be applied to the feature vector.

In this domain, there are two types of perturbations that can be applied to a feature vector. Feature addition is where a value in a feature vector is modified from 0 to 1. %
In the problem-space, an attacker can achieve this in different ways, such as adding dead code so that the feature vector representing the software changes to perform this perturbation, or by using opaque predicates \cite{moser2007limits, pierazzi2020problemspace}. This has proved to work well to create adversarial malware, for instance in Windows~\cite{demetrio2021functionality}. It should be noted that analysis of the call graph by a defender may be able to detect the dead code. Meanwhile, feature removal is where a value in a feature vector is modified from 1 to 0. This is a more complex operation, as there is a chance of removing features affecting functionality \cite{li2021framework, 8171381, li2020enhancing, pierazzi2020problemspace}. For Android apps, an attacker cannot remove features from the manifest file nor intent filter, and component names must be consistently named. Furthermore, the S6 feature family of DREBIN is dependent upon other feature families and cannot be removed. Therefore, the opportunities for feature removal lie in areas such as rewriting dexcode to achieve the same functionality, encrypting system/API calls and network addresses. For example, obfuscating API calls would allow those \emph{features} to be removed (since they would then count as new features) even though the functionality would remain \cite{li2020enhancing, pierazzi2020problemspace}. 

For each dataset, it is necessary to consider the allowed perturbations for generating adversarial examples. We achieve this by consulting with industry documentation and previous work \cite{li2021framework, 8171381, li2020enhancing, al2018adversarial, pierazzi2020problemspace, labaca2021universal}. We find that DREBIN allows for both feature addition and removal, with Table~\ref{table:allowedperturbationsdrebin} providing a summary of the allowed perturbations for each of the feature families \cite{li2021framework, 8171381}. Meanwhile, for SLEIPNIR, we can only perform feature addition because of the encapsulation performed by the feature extraction mechanism of LIEF when the dataset was originally developed.

\begin{table}[!htbp]
\centering
\scalebox{0.8}{
\begin{tabular}{llcc|c}
    \hline
     & Feature families & \multicolumn{1}{l}{Addition} & \multicolumn{1}{l}{Removal}  & \multicolumn{1}{l}{\makecell[l]{Usage in AE\\ generation}}\\ \hline
    \multirow{4}{*}{manifest} & S1 Hardware & \cmark & \xmark & 0.1\%\\
     & S2 Requested permissions & \cmark & \xmark & 1\% \\
     & S3 Application components & \cmark & \cmark & 41.4\% \\
     & S4 Intents & \cmark & \xmark & 1.8\% \\ \hline
    \multirow{4}{*}{dexcode} & S5 Restricted API Calls & \cmark & \cmark & 0.5\% \\
     & S6 Used permission & \xmark & \xmark & 0\% \\
     & S7 Suspicious API calls & \cmark & \cmark & 0.5\% \\
     & S8 Network addresses & \cmark & \cmark & 54.7\% \\ \hline
    \end{tabular}
}
\caption{Allowed perturbations for DREBIN feature families. The final column presents the frequency of each feature family in their use for perturbing input samples for generating adversarial examples (AEs) following the procedure in Section~\ref{sec:crafting}.}
\label{table:allowedperturbationsdrebin}
\end{table}

To ensure that perturbations remain valid, we include a verification step in our attack pipeline to monitor perturbations applied to a feature vector. Firstly, attacks are applied to malware samples to generate adversarial examples without any limitations. Then, because the attacks we use produce continuous feature vectors, their values are rounded to the nearest integer (i.e., 0 or 1) to represent the presence or absence of that feature. Each adversarial example is then inspected for prohibited perturbations, which are reversed. For example, if a feature from the S2 family is removed by an attack for DREBIN, then the original value is restored as it is not allowed to be removed. As this process can change back features used to cross the decision boundary (e.g., from malware to benign), we then ensure that the adversarial example is still adversarial by testing it on the model.

In Section~\ref{sec:crafting}, we provide a description of the procedure used for generating adversarial examples. Based on these generated adversarial examples, there are certain practical chracteristics that are interesting to study. For example, for DREBIN, Table~\ref{table:allowedperturbationsdrebin} presents a summary of the feature families and whether features from those families can be added or removed to generate adversarial examples. This table additionally shows an interesting analysis of the frequency of each family in how they contribute to the generation for adversarial examples. We find that features from the the families S8 (network addresses) and S3 (application components)  are most frequently perturbed in order to generate adversarial examples. This is further reflected in Table~\ref{table:featurefamiliesusedmostdrebin}, which presents the top 10 most perturbed features used for generating adversarial examples in the following section. Here, we once again observe that features from the S3 and S8 are most commonly perturbed to have input samples cross the decision boundary.

\begin{table}[!htbp]
\centering
\scalebox{0.85}{
\begin{tabular}{ll}
\hline
Actual Feature & Feature Family \\
\hline
activity::.optionsapp & S3\\
activity::.resend & S3 \\
url::45667 & S8 \\
url::34992 & S8 \\
activity::.mservice & S3 \\
activity::11628 & S3 \\
activity::.6244 & S3 \\
service\_receiver::.calllistener & S3 \\
call::getsystemservice & S7 \\
activity::.rebootbutton & S3 \\
\hline
\end{tabular}
}
\caption{Top 10 features that are perturbed most in order to generate adversarial examples for DREBIN. We have obfuscated some activities and network addresses.}
\label{table:featurefamiliesusedmostdrebin}
\end{table}

Although a similar analysis can be conducted for SLEIPNIR, the original dataset does not provide the names of the API calls that form the features. Hence, it is not possible to report the names of the API calls that are most frequently used to generate adversarial examples.

\subsection{Procedure for Generating Adversarial Examples}
\label{sec:crafting}
We generate adversarial examples in the feature-space like previous work \cite{li2020enhancing, al2018adversarial, grosse2017adversarial, demontis2017yes}. When doing this, we ensure that feature vectors remain discrete and that malicious functionality is preserved by limiting the set of allowed perturbations that can be applied to the feature vector. This ensures that adversarial examples remain close to realistic and functional malware, without the need for testing in a sandbox environment.

As detailed in Section \ref{sec:applicationandthreatmodel}, our threat model mainly consists of a gray-box scenario where the attacker's knowledge is limited~\cite{papernot2018sok, laskov2014practical, santana2021detecting, biggio2018wild}, so we focus in this section on describing the process we follow for this. We also consider a black-box scenario, but this is described in detail in Section \ref{sec:blackboxattack}. In particular, for the gray-box scenario, attackers have access to the same training data as the target model and have knowledge of the feature representation.  %
Therefore, to simulate this scenario, we construct four substitute vanilla models using the training data: a decision tree (DT), neural network (NN), random forest (RF) and support vector machine (SVM) (see Appendix~\ref{appendix:architecutres} for model architectures). It is well established that using models with architectures different from the target model can be used to evade it \cite{szegedy2013intriguing}. Therefore, we apply the attacks listed in Table~\ref{table:attackimps} against these substitute models to generate adversarial examples. We can apply white-box attacks to the substitute models because we have access to their gradient information. An overview of the procedure for generating adversarial examples is provided:

\begin{enumerate}
    \item With an input malware sample and an (applicable) substitute model (\(S\)), an attack (\(\tau\)) is performed, to generate an adversarial example. The malware samples are those from our test set.
    \item If the generated feature vector is continuous, the values within are rounded to the nearest integer (i.e., 0 or 1), in order to restore it to a discrete vector.
    \item We then verify that all perturbations are valid according to the dataset. Any invalid perturbations are reverted, to offer a lower bound of functionality preservation within the feature-space, similar to prior work \cite{severi2021explanation, grosse2017adversarial, li2021framework}.
    \item The adversarial example is then evaluated to ensure it is still adversarial. The substitute model \(S\) makes a prediction for the original input sample and the adversarial example; a difference between them indicates that the adversarial example has crossed the decision boundary. If so, it is added to the set of adversarial examples generated by attack \(\tau\) against \(S\), which is represented by \(\Omega_{\tau, S}\). 
    \item The sets of adversarial examples can be tested on StratDef, following the steps in the next sections.
\end{enumerate}

\noindent{This} process results in 4608 unique adversarial examples for DREBIN and 5640 for SLEIPNIR. \(\Omega_{DREBIN}\) and \(\Omega_{SLEIPNIR}\) are the sets of adversarial examples for DREBIN and SLEIPNIR respectively. The procedure is performed by the defender and the attacker independently. Different attacker profiles are then constructed as described in Section \ref{sec:attackerprofiles} based on the generated adversarial examples. The defender uses the generated adversarial examples (together with the training and validation data) as part of the process described in the next sections. %

\begin{table}[!htbp]
\centering
\scalebox{0.85}{
    \begin{tabular}{lll} 
    \hline
    Attack Name & Applicable Model Families \\
    \hline
    Basic Iterative Method (A) \cite{kurakin2016adversarial} & NN \\
    Basic Iterative Method (B) \cite{kurakin2016adversarial} & NN \\
    Boundary Attack \cite{brendel2017decision} & DT, NN, RF, SVM \\
    Carlini Attack \cite{carlini2017towards} & NN, SVM \\
    Decision Tree Attack \cite{grosse2017statistical} & DT \\
    Deepfool Attack \cite{moosavi2016deepfool} & NN, SVM \\
    Fast Gradient Sign Method \cite{goodfellow2014explaining, dhaliwal2018gradient} & NN, SVM \\
    HopSkipJump Attack \cite{chen2020hopskipjumpattack} & DT, NN, RF, SVM \\
    Jacobian Saliency Map Approach \cite{papernot2016limitations}& NN, SVM \\
    Project Gradient Descent\cite{madry2017towards} & NN, SVM \\
    Support Vector Machine Attack  \cite{grosse2017statistical} & SVM \\
\hline
    \end{tabular}
}
\caption{Attacks used to generate adversarial examples. Some attacks can only be applied to certain model families.}
\label{table:attackimps}
\end{table}

\subsection{Modelling Gray-Box Attacker Profiles}
\label{sec:attackerprofiles}
After generating each set of adversarial examples \(\Omega_{\tau, S}\) as detailed in the previous section, we assign each set to different attacker profiles, according to Table~\ref{table:attackerprofiles}. The aim of this is to simulate and evaluate StratDef's performance against different types of attackers, similar to prior work \cite{carlini2017adversarial, carlini2019evaluating, biggio2018wild, 217486, severi2021explanation, pierazzi2020problemspace}. %

When modeling attacker profiles, we ensure that the strongest attacker only uses the sets of adversarial examples with higher normalized evasion rates\footnote{Recall that the normalized evasion rate encompasses the number of adversarial examples in a particular set and their overall evasiveness (see Section~\ref{sec:howtomovedesc}).} against each model \(F \in \Sigma_{\gamma}\). Weaker attackers use those sets with lower normalized evasion rates. Additionally, stronger attackers can \emph{observe transferability}. If an attacker cannot observe transferability, then when assigning them a set of adversarial examples, we only consider the normalized evasion rate against the substitute model \(S\), which is the original applicable substitute %
model, and not against %
models in \(\Sigma_{\gamma}\), which could be higher due to transferability. %

Once an attacker has been assigned sets of adversarial examples, the sets are aggregated into a single set so that each attacker has a selection of adversarial examples to model and represent their capability. Using these, we create datasets to represent different attack intensities, denoted by \(\alpha\), for each attacker. The attack intensity represents the proportion of adversarial examples in the dataset (i.e., adversarial queries made by attackers). Each dataset corresponds to a value of \(\alpha \in\left[0,1\right]\) with increments of 0.1. For example, at \(\alpha = 0.1\), 10\% of the dataset consists of adversarial examples. The remaining 90\% consists of an equal number of benign and \emph{non-adversarial} malware samples from the test set described in Section~\ref{sec:experimentalsetupdatasets}. This would therefore represent a system condition (or operating environment) with a lower attack or adversarial intensity as compared with \(\alpha=0.9\), for example. The \emph{pooling procedure} in Table~\ref{table:attackerprofiles} governs how the adversarial examples for the \(\alpha\) datasets are chosen. For the strong attacker, the construction of these datasets gives preference to more evasive adversarial examples from their aggregated set. That is, to construct the \(\alpha\) datasets, adversarial examples that are from more evasive sets have a higher chance of being chosen. This is unlike for other attackers, where the adversarial examples are chosen randomly from their aggregated set. Meanwhile, the universal attacker only gets assigned adversarial examples with %
universal adversarial perturbations (UAPs) \cite{moosavi2017universal, labaca2021universal}. In total, there are 1,541 such universal examples for DREBIN (\(UAP_{DREBIN}\)) and 2,217 for SLEIPNIR (\(UAP_{SLEIPNIR}\)) (see Section~\ref{sec:dealingwithuaps} later).

\begin{table}[!htbp]
\centering
\scalebox{0.85}{
\begin{tabular}{lllll}
\hline
Attacker & Strength & \makecell[l]{Observe\\ Transferability} & \makecell[l]{Pooling\\ Procedure}\\
\hline
Weak  & \(R_{\tau, S, F} \leq 40\)  & \xmark & Random\\
\hline
Medium &  \(40 \leq R_{\tau, S, F} \leq 80\)  & \cmark & Random \\
\hline
Strong & \(80 \leq R_{\tau, S, F}\)  & \cmark & 
\makecell[l]{Weighted\\ based on \(R_{\tau, S, F}\)}\\
\hline
Random & Any & \cmark & Random\\
\hline
Universal & UAPs only & \cmark & Random\\
\hline

\end{tabular}
}

\caption{Profiles of different gray-box attackers who interact with StratDef. For the black-box attacker, see Section~\ref{sec:blackboxattack}.}
\label{table:attackerprofiles}
\end{table}

The datasets generated for each attacker are also used by the defender for developing strategies (as per Section~\ref{sec:whattomovedescpartb}) and evaluating the performance of each defense by simulating attackers with different levels of adversarial queries. We next present our experimental evaluation and results.

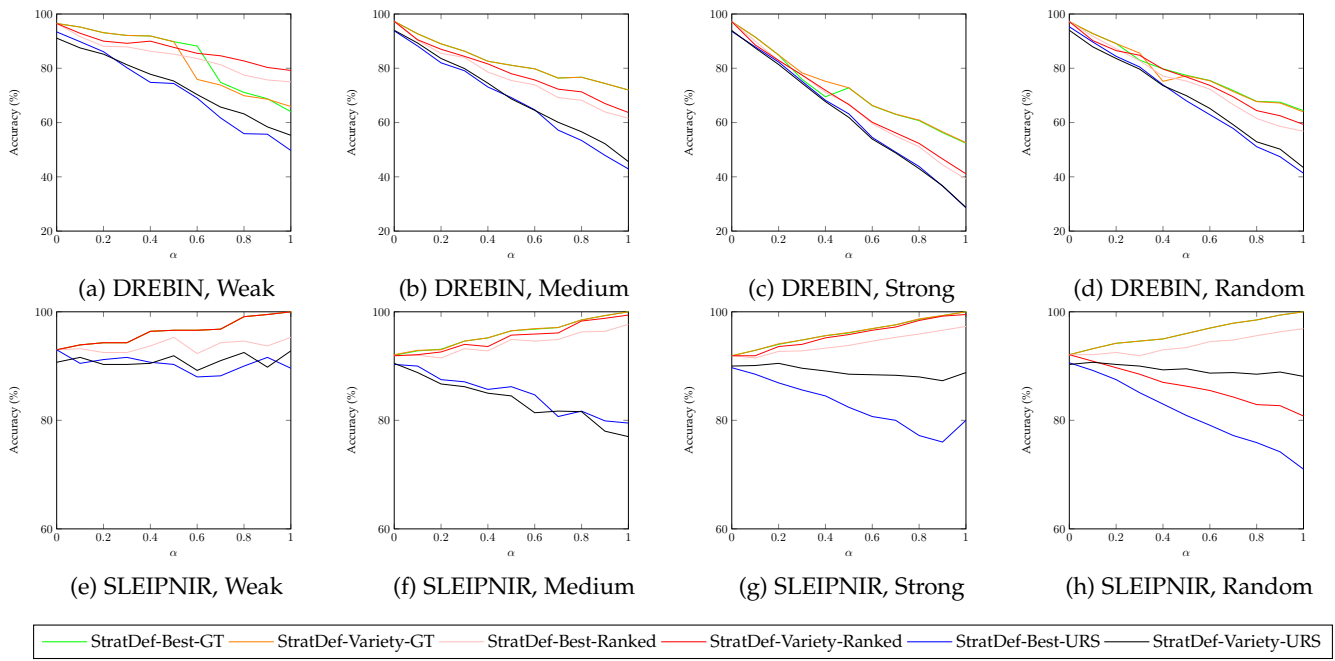
\begin{figure*}[!b]
\centering
\begin{subfigure}[b]{0.25\textwidth}
    \begin{tikzpicture}[scale=0.45]
        \begin{axis}[
            xlabel={\(\alpha\)},
            ylabel={Accuracy (\%)}, 
            xmin=0.0, xmax=1.0,
            ymin=20, ymax=100,
            xtick={0.0,0.2,0.4,0.6,0.8,1.0},
            ytick={20,40,60,80,100},
            ymajorgrids=false,
            legend pos=south west,
            legend style={nodes={scale=0.7, transform shape}},
            width=8.5cm,
            height=8cm,
            legend to name=stratdeflegendmainfigure
        ]
        
        \addplot[
            color=green,
            ]
            coordinates {
            (0,96.5)(0.1,95.2)(0.2,93.1)(0.3,92.1)(0.4,91.9)(0.5,89.8)(0.6,88.2)(0.7,74.8)(0.8,71.1)(0.9,68.8)(1,64)
            };
        \addlegendentry{StratDef-Best-GT}

        \addplot[
            color=orange,
            ]
            coordinates {
            (0,96.5)(0.1,95.2)(0.2,93.1)(0.3,92.1)(0.4,91.9)(0.5,89.8)(0.6,75.9)(0.7,73.8)(0.8,69.9)(0.9,68.6)(1,65.9)
            };
        \addlegendentry{StratDef-Variety-GT}
        
         \addplot[
            color=pink,
            ]
            coordinates {
             (0,96.5)(0.1,91.7)(0.2,88.1)(0.3,87.9)(0.4,86.3)(0.5,85.2)(0.6,83.6)(0.7,81.3)(0.8,77.5)(0.9,75.7)(1,75)
            };
        \addlegendentry{StratDef-Best-Ranked}

         \addplot[
            color=red,
            ]
            coordinates {
           (0,96.5)(0.1,92.9)(0.2,90)(0.3,89.2)(0.4,90)(0.5,87.7)(0.6,85.5)(0.7,84.6)(0.8,82.7)(0.9,80.3)(1,79.2)
            };
        \addlegendentry{StratDef-Variety-Ranked}

         \addplot[
            color=blue,
            ]
            coordinates {
            (0,93.4)(0.1,89.8)(0.2,86.1)(0.3,80.2)(0.4,74.8)(0.5,74.4)(0.6,69)(0.7,61.7)(0.8,55.9)(0.9,55.7)(1,49.7)
            };
        \addlegendentry{StratDef-Best-URS}
        
         \addplot[
            color=black,
            ]
            coordinates {
            (0,91.1)(0.1,87.5)(0.2,85.2)(0.3,81.3)(0.4,77.8)(0.5,75.3)(0.6,70.3)(0.7,65.7)(0.8,63.2)(0.9,58.4)(1,55.3)
            };
        \addlegendentry{StratDef-Variety-URS}

        \end{axis}
    \end{tikzpicture}
\caption{DREBIN, Weak}
\end{subfigure}
\hskip -1ex
\begin{subfigure}[b]{0.25\textwidth}
    \begin{tikzpicture}[scale=0.45]
        \begin{axis}[
            xlabel={\(\alpha\)},
            ylabel={Accuracy (\%)}, 
            xmin=0.0, xmax=1.0,
            ymin=20, ymax=100,
            xtick={0.0,0.2,0.4,0.6,0.8,1.0},
            ytick={20,40,60,80,100},
            ymajorgrids=false,
            legend pos=south west,
            legend style={nodes={scale=0.7, transform shape}},
            width=8.5cm,
            height=8cm,
            legend to name=stratdeflegendmainfigure
        ]
        
        \addplot[
            color=green,
            ]
            coordinates {
            (0,97.3)(0.1,92.7)(0.2,89)(0.3,86.3)(0.4,82.6)(0.5,81.1)(0.6,79.8)(0.7,76.4)(0.8,76.7)(0.9,74.4)(1,72)
            };
        \addlegendentry{StratDef-Best-GT}

        \addplot[
            color=orange,
            ]
            coordinates {
            (0,97.3)(0.1,92.7)(0.2,89)(0.3,86.3)(0.4,82.6)(0.5,81.1)(0.6,79.8)(0.7,76.4)(0.8,76.7)(0.9,74.4)(1,72)
            };
        \addlegendentry{StratDef-Variety-GT}
        
         \addplot[
            color=pink,
            ]
            coordinates {
             (0,97.3)(0.1,90.3)(0.2,85.5)(0.3,84)(0.4,78.6)(0.5,75.5)(0.6,73.9)(0.7,69.2)(0.8,68.2)(0.9,63.9)(1,61.6)
            };
        \addlegendentry{StratDef-Best-Ranked}

         \addplot[
            color=red,
            ]
            coordinates {
           (0,97.3)(0.1,90.5)(0.2,87)(0.3,84.4)(0.4,81.6)(0.5,78)(0.6,75.7)(0.7,72.3)(0.8,71.3)(0.9,67)(1,63.7)
            };
        \addlegendentry{StratDef-Variety-Ranked}

         \addplot[
            color=blue,
            ]
            coordinates {
            (0,93.8)(0.1,88.2)(0.2,82)(0.3,79.1)(0.4,73.1)(0.5,69.2)(0.6,64.7)(0.7,57.2)(0.8,53.4)(0.9,47.9)(1,42.9)
            };
        \addlegendentry{StratDef-Best-URS}
        
         \addplot[
            color=black,
            ]
            coordinates {
            (0,94)(0.1,89.3)(0.2,83.6)(0.3,79.9)(0.4,74.5)(0.5,68.7)(0.6,64.5)(0.7,60.1)(0.8,56.6)(0.9,52.2)(1,45.6)
            };
        \addlegendentry{StratDef-Variety-URS}

        \end{axis}
    \end{tikzpicture}
\caption{DREBIN, Medium}
\label{figure:drebinmedium}
\end{subfigure}
\hskip -1ex
\begin{subfigure}[b]{0.25\textwidth}
    \begin{tikzpicture}[scale=0.45]
        \begin{axis}[
            xlabel={\(\alpha\)},
            ylabel={Accuracy (\%)}, 
            xmin=0.0, xmax=1.0,
            ymin=20, ymax=100,
            xtick={0.0,0.2,0.4,0.6,0.8,1.0},
            ytick={20,40,60,80,100},
            ymajorgrids=false,
            legend pos=south west,
            legend style={nodes={scale=0.7, transform shape}},
            width=8.5cm,
            height=8cm,
            legend to name=stratdeflegendmainfigure
        ]
        
        \addplot[
            color=green,
            ]
            coordinates {
            (0,97.2)(0.1,91.5)(0.2,84.9)(0.3,75.9)(0.4,69.5)(0.5,72.8)(0.6,66.2)(0.7,63)(0.8,60.7)(0.9,56.3)(1,52.4)
            };
        \addlegendentry{StratDef-Best-GT}

        \addplot[
            color=orange,
            ]
            coordinates {
            (0,97.2)(0.1,91.5)(0.2,84.9)(0.3,78.3)(0.4,75.2)(0.5,72.8)(0.6,66.3)(0.7,63.1)(0.8,60.9)(0.9,56.6)(1,52.6)
            };
        \addlegendentry{StratDef-Variety-GT}
        
         \addplot[
            color=pink,
            ]
            coordinates {
             (0,97.2)(0.1,90)(0.2,82.8)(0.3,77.6)(0.4,71.1)(0.5,66.9)(0.6,59.6)(0.7,55)(0.8,51.1)(0.9,44.4)(1,39.2)
            };
        \addlegendentry{StratDef-Best-Ranked}

         \addplot[
            color=red,
            ]
            coordinates {
           (0,97.2)(0.1,88.6)(0.2,82.9)(0.3,77.5)(0.4,71.9)(0.5,66.6)(0.6,60.1)(0.7,56.2)(0.8,52.3)(0.9,46.6)(1,41.1)
            };
        \addlegendentry{StratDef-Variety-Ranked}

         \addplot[
            color=blue,
            ]
            coordinates {
            (0,93.6)(0.1,87.9)(0.2,82.3)(0.3,75.2)(0.4,68.2)(0.5,63.2)(0.6,54.5)(0.7,49.1)(0.8,43.8)(0.9,36.6)(1,28.6)
            };
        \addlegendentry{StratDef-Best-URS}
        
         \addplot[
            color=black,
            ]
            coordinates {
            (0,93.9)(0.1,87.5)(0.2,81.4)(0.3,74.5)(0.4,67.7)(0.5,61.9)(0.6,53.9)(0.7,48.8)(0.8,43)(0.9,36.7)(1,28.7)
            };
        \addlegendentry{StratDef-Variety-URS}

        \end{axis}
    \end{tikzpicture}
\caption{DREBIN, Strong}
\end{subfigure}
\hskip -1ex
\begin{subfigure}[b]{0.25\textwidth}
    \begin{tikzpicture}[scale=0.45]
        \begin{axis}[
            xlabel={\(\alpha\)},
            ylabel={Accuracy (\%)}, 
            xmin=0.0, xmax=1.0,
            ymin=20, ymax=100,
            xtick={0.0,0.2,0.4,0.6,0.8,1.0},
            ytick={20,40,60,80,100},
            ymajorgrids=false,
            legend pos=south west,
            legend style={nodes={scale=0.7, transform shape}},
            width=8.5cm,
            height=8cm,
            legend to name=stratdeflegendmainfigure
        ]
        
        \addplot[
            color=green,
            ]
            coordinates {
            (0,97.2)(0.1,92.8)(0.2,89.1)(0.3,83)(0.4,79.8)(0.5,77.4)(0.6,75.5)(0.7,71.8)(0.8,67.8)(0.9,67.5)(1,64.4)
            };
        \addlegendentry{StratDef-Best-GT}

        \addplot[
            color=orange,
            ]
            coordinates {
            (0,97.2)(0.1,92.8)(0.2,89.1)(0.3,85.6)(0.4,75.2)(0.5,77.2)(0.6,75.4)(0.7,71.4)(0.8,67.7)(0.9,67.2)(1,63.9)
            };
        \addlegendentry{StratDef-Variety-GT}
        
         \addplot[
            color=pink,
            ]
            coordinates {
             (0,97.2)(0.1,91.8)(0.2,87.2)(0.3,82.8)(0.4,77.1)(0.5,75.3)(0.6,72.3)(0.7,66.6)(0.8,61.5)(0.9,58.6)(1,56.8)
            };
        \addlegendentry{StratDef-Best-Ranked}

         \addplot[
            color=red,
            ]
            coordinates {
           (0,97.2)(0.1,90.2)(0.2,86.5)(0.3,84.8)(0.4,79.7)(0.5,76.8)(0.6,73.7)(0.7,69.5)(0.8,64.4)(0.9,62.5)(1,59.3)
            };
        \addlegendentry{StratDef-Variety-Ranked}

         \addplot[
            color=blue,
            ]
            coordinates {
            (0,95.3)(0.1,89.7)(0.2,84.4)(0.3,80.5)(0.4,73.9)(0.5,68)(0.6,62.9)(0.7,57.8)(0.8,51.1)(0.9,47.4)(1,41.3)
            };
        \addlegendentry{StratDef-Best-URS}
        
         \addplot[
            color=black,
            ]
            coordinates {
            (0,94)(0.1,87.9)(0.2,83.6)(0.3,79.7)(0.4,73.6)(0.5,69.9)(0.6,65.2)(0.7,59.2)(0.8,52.9)(0.9,50.2)(1,43.4)
            };
        \addlegendentry{StratDef-Variety-URS}

        \end{axis}
    \end{tikzpicture}
\caption{DREBIN, Random}
\end{subfigure}

\begin{subfigure}[b]{0.25\textwidth}
    \begin{tikzpicture}[scale=0.45]
        \begin{axis}[
            xlabel={\(\alpha\)},
            ylabel={Accuracy (\%)}, 
            xmin=0.0, xmax=1.0,
            ymin=60, ymax=100,
            xtick={0.0,0.2,0.4,0.6,0.8,1.0},
            ytick={60,80,100},
            ymajorgrids=false,
            legend pos=south west,
            legend style={nodes={scale=0.7, transform shape}},
            width=8.5cm,
            height=8cm,
            legend to name=stratdeflegendmainfigure
        ]
        
        \addplot[
            color=green,
            ]
            coordinates {
            (0,93)(0.1,93.9)(0.2,94.3)(0.3,94.3)(0.4,96.4)(0.5,96.6)(0.6,96.6)(0.7,96.8)(0.8,99.1)(0.9,99.5)(1,100)
            };
        \addlegendentry{StratDef-Best-GT}

        \addplot[
            color=orange,
            ]
            coordinates {
            (0,93)(0.1,93.9)(0.2,94.3)(0.3,94.3)(0.4,96.4)(0.5,96.6)(0.6,96.6)(0.7,96.8)(0.8,99.1)(0.9,99.5)(1,100)
            };
        \addlegendentry{StratDef-Variety-GT}
        
         \addplot[
            color=pink,
            ]
            coordinates {
             (0,93)(0.1,93.2)(0.2,92.5)(0.3,92.5)(0.4,93.7)(0.5,95.3)(0.6,92.3)(0.7,94.3)(0.8,94.6)(0.9,93.7)(1,95.3)
            };
        \addlegendentry{StratDef-Best-Ranked}

         \addplot[
            color=red,
            ]
            coordinates {
           (0,93)(0.1,93.9)(0.2,94.3)(0.3,94.3)(0.4,96.4)(0.5,96.6)(0.6,96.6)(0.7,96.8)(0.8,99.1)(0.9,99.5)(1,100)
            };
        \addlegendentry{StratDef-Variety-Ranked}

         \addplot[
            color=blue,
            ]
            coordinates {
            (0,93)(0.1,90.5)(0.2,91.2)(0.3,91.6)(0.4,90.7)(0.5,90.3)(0.6,88)(0.7,88.2)(0.8,90)(0.9,91.6)(1,89.6)
            };
        \addlegendentry{StratDef-Best-URS}
        
         \addplot[
            color=black,
            ]
            coordinates {
            (0,90.7)(0.1,91.6)(0.2,90.3)(0.3,90.3)(0.4,90.5)(0.5,91.9)(0.6,89.2)(0.7,91)(0.8,92.5)(0.9,89.8)(1,92.8)
            };
        \addlegendentry{StratDef-Variety-URS}

        \end{axis}
    \end{tikzpicture}
\caption{SLEIPNIR, Weak}
\label{figure:sleipnirweak}
\end{subfigure}
\hskip -1ex
\begin{subfigure}[b]{0.25\textwidth}
    \begin{tikzpicture}[scale=0.45]
        \begin{axis}[
            xlabel={\(\alpha\)},
            ylabel={Accuracy (\%)}, 
            xmin=0.0, xmax=1.0,
            ymin=60, ymax=100,
            xtick={0.0,0.2,0.4,0.6,0.8,1.0},
            ytick={60,80,100},
            ymajorgrids=false,
            legend pos=south west,
            legend style={nodes={scale=0.7, transform shape}},
            width=8.5cm,
            height=8cm,
            legend to name=stratdeflegendmainfigure
        ]
        
        \addplot[
            color=green,
            ]
            coordinates {
            (0,92)(0.1,92.8)(0.2,93.1)(0.3,94.6)(0.4,95.2)(0.5,96.5)(0.6,96.8)(0.7,97.1)(0.8,98.5)(0.9,99.3)(1,100)
            };
        \addlegendentry{StratDef-Best-GT}

        \addplot[
            color=orange,
            ]
            coordinates {
            (0,92.1)(0.1,92.9)(0.2,93)(0.3,94.6)(0.4,95.2)(0.5,96.5)(0.6,96.9)(0.7,97.1)(0.8,98.5)(0.9,99.3)(1,100)
            };
        \addlegendentry{StratDef-Variety-GT}
        
         \addplot[
            color=pink,
            ]
            coordinates {
             (0,91.9)(0.1,92)(0.2,91.5)(0.3,93.2)(0.4,92.8)(0.5,94.9)(0.6,94.6)(0.7,94.9)(0.8,96.3)(0.9,96.4)(1,97.7)
            };
        \addlegendentry{StratDef-Best-Ranked}

         \addplot[
            color=red,
            ]
            coordinates {
           (0,91.9)(0.1,92.1)(0.2,92.6)(0.3,94)(0.4,93.6)(0.5,95.7)(0.6,95.9)(0.7,96.1)(0.8,98.3)(0.9,98.8)(1,99.4)
            };
        \addlegendentry{StratDef-Variety-Ranked}

         \addplot[
            color=blue,
            ]
            coordinates {
            (0,90.3)(0.1,90)(0.2,87.5)(0.3,87.1)(0.4,85.7)(0.5,86.2)(0.6,84.7)(0.7,80.7)(0.8,81.7)(0.9,79.9)(1,79.5)
            };
        \addlegendentry{StratDef-Best-URS}
        
         \addplot[
            color=black,
            ]
            coordinates {
            (0,90.5)(0.1,88.8)(0.2,86.7)(0.3,86.2)(0.4,85)(0.5,84.5)(0.6,81.4)(0.7,81.7)(0.8,81.6)(0.9,78)(1,77)
            };
        \addlegendentry{StratDef-Variety-URS}

        \end{axis}
    \end{tikzpicture}
\caption{SLEIPNIR, Medium}
\end{subfigure}
\hskip -1ex
\begin{subfigure}[b]{0.25\textwidth}
    \begin{tikzpicture}[scale=0.45]
        \begin{axis}[
            xlabel={\(\alpha\)},
            ylabel={Accuracy (\%)}, 
            xmin=0.0, xmax=1.0,
            ymin=60, ymax=100,
            xtick={0.0,0.2,0.4,0.6,0.8,1.0},
            ytick={60,80,100},
            ymajorgrids=false,
            legend pos=south west,
            legend style={nodes={scale=0.7, transform shape}},
            width=8.5cm,
            height=8cm,
            legend to name=stratdeflegendmainfigure
        ]
        
        \addplot[
            color=green,
            ]
            coordinates {
            (0,91.9)(0.1,92.9)(0.2,94)(0.3,94.8)(0.4,95.6)(0.5,96.1)(0.6,96.9)(0.7,97.6)(0.8,98.6)(0.9,99.3)(1,100)
            };
        \addlegendentry{StratDef-Best-GT}

        \addplot[
            color=orange,
            ]
            coordinates {
            (0,91.9)(0.1,92.9)(0.2,94.1)(0.3,94.8)(0.4,95.6)(0.5,96.2)(0.6,96.9)(0.7,97.6)(0.8,98.7)(0.9,99.3)(1,100)
            };
        \addlegendentry{StratDef-Variety-GT}
        
         \addplot[
            color=pink,
            ]
            coordinates {
             (0,91.9)(0.1,91.5)(0.2,92.7)(0.3,92.8)(0.4,93.3)(0.5,93.8)(0.6,94.6)(0.7,95.3)(0.8,95.9)(0.9,96.6)(1,97.3)
            };
        \addlegendentry{StratDef-Best-Ranked}

         \addplot[
            color=red,
            ]
            coordinates {
           (0,91.9)(0.1,91.9)(0.2,93.6)(0.3,94)(0.4,95.2)(0.5,95.8)(0.6,96.6)(0.7,97.2)(0.8,98.4)(0.9,99.2)(1,99.5)
            };
        \addlegendentry{StratDef-Variety-Ranked}

         \addplot[
            color=blue,
            ]
            coordinates {
            (0,89.7)(0.1,88.5)(0.2,86.9)(0.3,85.6)(0.4,84.5)(0.5,82.4)(0.6,80.7)(0.7,80)(0.8,77.2)(0.9,76)(1,80)
            };
        \addlegendentry{StratDef-Best-URS}
        
         \addplot[
            color=black,
            ]
            coordinates {
            (0,90)(0.1,90.1)(0.2,90.5)(0.3,89.6)(0.4,89.1)(0.5,88.5)(0.6,88.4)(0.7,88.3)(0.8,88)(0.9,87.3)(1,88.8)
            };
        \addlegendentry{StratDef-Variety-URS}

        \end{axis}
    \end{tikzpicture}
\caption{SLEIPNIR, Strong}
\end{subfigure}
\hskip -1ex
\begin{subfigure}[b]{0.25\textwidth}
    \begin{tikzpicture}[scale=0.45]
        \begin{axis}[
            xlabel={\(\alpha\)},
            ylabel={Accuracy (\%)}, 
            xmin=0.0, xmax=1.0,
            ymin=60, ymax=100,
            xtick={0.0,0.2,0.4,0.6,0.8,1.0},
            ytick={60,80,100},
            ymajorgrids=false,
            legend pos=south west,
            legend style={nodes={scale=0.7, transform shape}},
            width=8.5cm,
            height=8cm,
            legend to name=stratdeflegendmainfigure,
            legend columns=6
        ]
        
        \addplot[
            color=green,
            ]
            coordinates {
            (0,92.1)(0.1,93.2)(0.2,94.2)(0.3,94.6)(0.4,95)(0.5,96)(0.6,97)(0.7,97.9)(0.8,98.5)(0.9,99.4)(1,100)
            };
        \addlegendentry{StratDef-Best-GT}

        \addplot[
            color=orange,
            ]
            coordinates {
            (0,92.1)(0.1,93.2)(0.2,94.2)(0.3,94.6)(0.4,95)(0.5,96)(0.6,97)(0.7,97.9)(0.8,98.5)(0.9,99.4)(1,100)
            };
        \addlegendentry{StratDef-Variety-GT}
        
         \addplot[
            color=pink,
            ]
            coordinates {
             (0,92.1)(0.1,92.1)(0.2,92.5)(0.3,91.9)(0.4,93)(0.5,93.4)(0.6,94.5)(0.7,94.8)(0.8,95.6)(0.9,96.3)(1,96.9)
            };
        \addlegendentry{StratDef-Best-Ranked}

         \addplot[
            color=red,
            ]
            coordinates {
           (0,92.1)(0.1,90.9)(0.2,89.7)(0.3,88.5)(0.4,87)(0.5,86.3)(0.6,85.5)(0.7,84.3)(0.8,82.9)(0.9,82.7)(1,80.8)
            };
        \addlegendentry{StratDef-Variety-Ranked}

         \addplot[
            color=blue,
            ]
            coordinates {
            (0,90.6)(0.1,89.2)(0.2,87.5)(0.3,85.1)(0.4,83)(0.5,80.9)(0.6,79.1)(0.7,77.2)(0.8,75.9)(0.9,74.2)(1,71)
            };
        \addlegendentry{StratDef-Best-URS}
        
         \addplot[
            color=black,
            ]
            coordinates {
            (0,90.3)(0.1,90.7)(0.2,90.3)(0.3,90)(0.4,89.3)(0.5,89.5)(0.6,88.7)(0.7,88.8)(0.8,88.5)(0.9,88.9)(1,88.1)
            };
        \addlegendentry{StratDef-Variety-URS}

        \end{axis}
    \end{tikzpicture}
\caption{SLEIPNIR, Random}
\label{figure:sleipnirrandom}
\end{subfigure}

\vspace{3mm}
\ref{stratdeflegendmainfigure}
\caption{Accuracy of different StratDef configurations against different attackers under varying attack intensities. In Figures~\ref{figure:drebinmedium}, \ref{figure:sleipnirweak} and \ref{figure:sleipnirrandom}, StratDef-Best-GT and StratDef-Variety-GT have the same performance because of identical strategies.}
\label{figure:mtdresults}
\end{figure*}

\section{Evaluation}
\label{sec:evaluation}
In this section, we present an evaluation across Android and Windows in the experimental setting described previously. In Section~\ref{sec:performanceourdefense}, we analyze the performance of StratDef under different threat levels. Then, we compare StratDef's performance with other defenses in Section~\ref{sec:comparisonother} and compare its efficiency with other defenses in  Section~\ref{sec:expefficieny} later. Following this, we evaluate how StratDef performs against Universal Adversarial Perturbations in Section~\ref{sec:dealingwithuaps}. Finally, we show how StratDef copes with a complete black-box attacker in Section~\ref{sec:blackboxattack}.

\subsection{Performance of StratDef}
\label{sec:performanceourdefense}
In this section, we present the results for the StratDef configurations against different attackers and attack intensities. In our discussion, we focus on the aggregate results but refer the reader to Appendix~\ref{appendix:extendedresults} for extended results. %

Figure~\ref{figure:mtdresults} demonstrates that as the threat level increases (with stronger attackers and higher intensities), there is a greater effect on the performance of StratDef. Notably, as the threat level increases, we observe decreases in accuracy in many cases, as one may expect. However, at the peak threat level, StratDef can achieve up to 52.6\% accuracy for DREBIN and 100\% accuracy for SLEIPNIR, with the highest average accuracy of 72.7\% for DREBIN and 96.2\% for SLEIPNIR across all attack intensities for all configurations. These results demonstrate that the adversarial examples for DREBIN are more evasive than those for SLEIPNIR, as indicated by StratDef's lower accuracy. This is likely because there are more perturbations that can be used to generate adversarial examples with DREBIN (i.e., feature addition \emph{and} removal). Hence, the attack surface is widened as there are more avenues for evasion leading to more effective attacks. In the case of SLEIPNIR, defenses have it easier to defend themselves. In fact, for SLEIPNIR, the weakest StratDef configuration (StratDef-Best-URS) only drops to \(\approx\) 80\% accuracy at the peak threat level. The weaker adversarial examples for SLEIPNIR can be attributed to more limitations in the perturbations that can be applied, therefore reducing the attack surface. 
Interestingly in some cases for SLEIPNIR, the performance of StratDef increases with the attack intensity ($\alpha$). Recall that as $\alpha$ increases, the model selections and strategies are designed to cope better with more hostile conditions. Hence, such configurations exhibit greater accuracy as they can identify weaker adversarial examples, especially when they are increasing in their number (which occurs as $\alpha$ increases) leading to improved performance. As the adversarial examples for SLEIPNIR are less evasive (as explained prior), this phenomenon is primarily observed for this dataset.

\begin{figure}[!htbp]
\centering
\begin{subfigure}[b]{0.25\textwidth}
\begin{tikzpicture}[scale=0.45]
    \begin{axis}[
        ybar,
        enlargelimits=0.15,
        ylabel={\%},
        ymax=100,
        symbolic x coords={Accuracy, F1, AUC, FPR},
        xtick=data,
        ytick={0,20,40,60,80,100},
        nodes near coords align={horizontal},
        legend pos=north east,
        legend style={nodes={scale=0.5, transform shape}},
        legend to name=avgmtdvalueslegend,
        legend columns=4,
        width=9cm,
        height=6cm
        ]
    \addplot coordinates {(Accuracy, 71.9) (F1, 78.5) (AUC, 78.6) (FPR, 7) };
    \addplot coordinates {(Accuracy, 72.7) (F1, 79.3) (AUC, 76.9) (FPR, 6.9) };
    \addplot coordinates {(Accuracy, 66.8) (F1, 73.4) (AUC, 87.6) (FPR, 4.8) };
    \addplot coordinates {(Accuracy, 67.4) (F1, 74.2) (AUC, 79.8) (FPR, 6.9) };
    
    \legend{Best, Variety, Best-Ranked, Variety-Ranked}
    \end{axis}
   
\end{tikzpicture}
\caption{DREBIN}
\end{subfigure}
\hskip -1ex
\begin{subfigure}[b]{0.22\textwidth}
\begin{tikzpicture}[scale=0.45]
    \begin{axis}[
        ybar,
        enlargelimits=0.15,
        ylabel={\%},
        ymax=100,
        symbolic x coords={Accuracy, F1, AUC, FPR},
        xtick=data,
        ytick={0,20,40,60,80,100},
        nodes near coords align={horizontal},
        legend pos=north east,
        legend style={nodes={scale=0.5, transform shape}},
        legend to name=avgmtdvalueslegend,
        legend columns=4,
        width=9cm,
        height=6cm
        ]
    \addplot coordinates {(Accuracy, 96.2) (F1, 96.9) (AUC, 99.2) (FPR, 8.6) };
    \addplot coordinates {(Accuracy, 96.2) (F1, 97) (AUC, 99.2) (FPR, 8.7) };
    \addplot coordinates {(Accuracy, 94.2) (F1, 95.6) (AUC, 97.9) (FPR, 10.3) };
    \addplot coordinates {(Accuracy, 95.8) (F1, 96.7) (AUC, 98.6) (FPR, 9.7) };
    
    \legend{StratDef-Best-GT, StratDef-Variety-GT, StratDef-Best-Ranked, StratDef-Variety-Ranked}
    \end{axis}
   
\end{tikzpicture}
\caption{SLEIPNIR}
\end{subfigure}

\ref{avgmtdvalueslegend}

\caption{Average values of metrics across attack intensities for StratDef configurations against the strong attacker.
}
\label{figure:avgmtdvalues}
\end{figure}
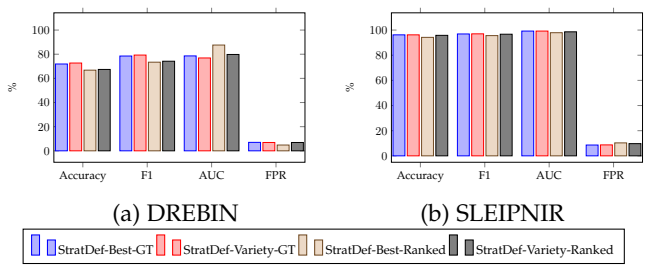

In terms of the model selection, the configurations using the \emph{Variety} model selection perform well at all threat levels. Recall that this model selection enforces diversity among the constituent models of the ensemble (see Section~\ref{sec:whattomovedescpartb}). The results show that the increased diversity in this model selection offsets the transferability of adversarial examples better than the configurations using the \emph{Best} model selection. This is further reflected, albeit marginally, across all metrics, as visible in Figure~\ref{figure:avgmtdvalues}. This figure shows the values of different metrics for StratDef configurations averaged across attack intensities for both datasets against the strong attacker. %

Meanwhile, regarding the optimizer, the \emph{game-theoretic (GT)} configurations offer slightly better performance for both datasets than the configurations using the \emph{strategic ranked optimizer (Ranked)}. However, a potential limitation of the GT configurations is that they switch between pure and mixed strategies, with adversarially-trained models featuring more often in the strategies. In fact, with the GT configurations, only up to 30\% of the model selection is used against the strong attacker, meaning that the majority of the model selection may never actually be used. It is important to understand that employing a StratDef configuration (or an MTD) that uses fewer models may increase the risk of an attacker discovering the profile and configuration of the deployed defense. However, due to transferability, using more models may open an avenue for greater evasion as there is a higher chance that a vulnerable constituent model is chosen at prediction-time. Therefore, an inherent trade-off exists between the number of models used in the ensemble and the robustness of the system. If a more diverse set of models is used to reduce the transferability, the attacker will be less successful.

The configurations using the strategic ranked optimizer only produce mixed strategies since this optimizer does not give complete preference to the strongest model. Intriguingly, despite using more models, these configurations offer similar performance to those that use the GT optimizer. This is especially visible in Figure~\ref{figure:avgmtdvalues}, where  we observe that the F1 and AUC tend to remain generally high, while the FPR tends to remain quite low, for the GT and Ranked configurations across the board. Overall, these results illustrate the capability of our strategic defense to produce correct predictions across a number of scenarios, especially adversarial ones, which is quite encouraging. In contrast, the poorer performance expected of the uniform random strategy (URS) approach can be seen in Figures~\ref{figure:mtdresults} and \ref{figure:avgmtdvalues}, which highlights the need for good movement strategies regardless of how sound the model selection itself is. Recall that under this configuration --- which is only included as a baseline --- a model is chosen randomly from the ensemble to serve a user's prediction, rather than with a strategic or heuristic approach. The results show that using a randomized approach for choosing models at prediction-time is no match for a game-theoretic or heuristically-driven strategy. For maximal robustness, only strategic or game-theoretic movement approaches are suitable.

\subsection{StratDef vs. Other Defenses}
\label{sec:comparisonother}

\begin{figure}[!h]
\centering
\begin{subfigure}[b]{0.25\textwidth}
    \begin{tikzpicture}[scale=0.45]
        \begin{axis}[
            xlabel={\(\alpha\)},
            ylabel={Accuracy (\%)}, 
            xmin=0.0, xmax=1.0,
            ymin=00, ymax=100,
            xtick={0.0,0.2,0.4,0.6,0.8,1.0},
            ytick={0,20,40,60,80,100},
            ymajorgrids=false,
            legend pos=north east,
            legend style={nodes={scale=0.5, transform shape}},
            legend to name=stratdefmtdcompareallsleipnirlegend,
            legend columns=3,
            cycle list name=color list,
            width=9cm,
            height=8cm
        ]
        
       \addplot
            coordinates {
            (0,90.3)(0.1,84.6)(0.2,79.4)(0.3,73.4)(0.4,67.3)(0.5,63.9)(0.6,55.9)(0.7,53.3)(0.8,47)(0.9,41.3)(1,34.4)
            };
        \addlegendentry{DT}
        \addplot
            coordinates {
            (0,96.8)(0.1,90.2)(0.2,83)(0.3,76.1)(0.4,68.2)(0.5,62.3)(0.6,54.2)(0.7,48.6)(0.8,42.4)(0.9,35)(1,26.4)
            };
        \addlegendentry{RF}
         \addplot
            coordinates {
             (0,95.8)(0.1,87.2)(0.2,78.5)(0.3,69.2)(0.4,60.6)(0.5,52.2)(0.6,42.9)(0.7,34.7)(0.8,25.9)(0.9,17.5)(1,7.6)
            };
        \addlegendentry{NN-DD}
         \addplot
            coordinates {
           (0,93.9)(0.1,86.9)(0.2,79.8)(0.3,72.4)(0.4,65)(0.5,57.9)(0.6,49.8)(0.7,44.2)(0.8,37.5)(0.9,30.2)(1,21.8)
            };
        \addlegendentry{SecSVM}
         \addplot
            coordinates {
            (0,90.4)(0.1,85.1)(0.2,80)(0.3,74)(0.4,68.6)(0.5,64.9)(0.6,57.1)(0.7,54.9)(0.8,49)(0.9,43.4)(1,37.3)
            };
        \addlegendentry{DT-RFN}
         \addplot
            coordinates {
            (0,90.3)(0.1,86.6)(0.2,83.1)(0.3,78.3)(0.4,75.2)(0.5,72.8)(0.6,66.3)(0.7,63.1)(0.8,60.9)(0.9,56.6)(1,52.6)
            };
        \addlegendentry{DT-AT-0.1}
         \addplot
            coordinates {
            (0,91.9)(0.1,85.7)(0.2,79.9)(0.3,73.9)(0.4,67.4)(0.5,61.9)(0.6,56)(0.7,51.2)(0.8,45.3)(0.9,39.2)(1,31.2)
            };
        \addlegendentry{NN-AT-0.25}
         \addplot
            coordinates {
            (0,97.2)(0.1,91.5)(0.2,84.9)(0.3,78.4)(0.4,71.8)(0.5,67.1)(0.6,59.1)(0.7,54.8)(0.8,49)(0.9,42.6)(1,33.8)
            };
        \addlegendentry{RF-AT-0.25}
        
         \addplot
            coordinates {
            (0,97.2)(0.1,91.5)(0.2,84.9)(0.3,78.3)(0.4,75.2)(0.5,72.8)(0.6,66.3)(0.7,63.1)(0.8,60.9)(0.9,56.6)(1,52.6)
            };
        \addlegendentry{StratDef-Variety-GT}
         \addplot
            coordinates {
            (0,97.2)(0.1,88.6)(0.2,82.9)(0.3,77.5)(0.4,71.9)(0.5,66.6)(0.6,60.1)(0.7,56.2)(0.8,52.3)(0.9,46.6)(1,41.1)
            };
        \addlegendentry{StratDef-Variety-Ranked}
        \addplot
            coordinates {
            (0,96.7)(0.1,90.3)(0.2,83.4)(0.3,76.8)(0.4,69.3)(0.5,64)(0.6,55.1)(0.7,50.7)(0.8,43.8)(0.9,37.4)(1,28.7)
            };
        \addlegendentry{Voting-Best-Majority}
        
         \addplot
            coordinates {
            (0,97)(0.1,90.7)(0.2,83.9)(0.3,77.4)(0.4,70.4)(0.5,65)(0.6,57.2)(0.7,51.7)(0.8,46.6)(0.9,38.9)(1,30.9)
            };
        \addlegendentry{Voting-Variety-Majority}
            
        \end{axis}
    \end{tikzpicture}
\caption{DREBIN}
\end{subfigure}
\hskip -1ex
\begin{subfigure}[b]{0.22\textwidth}
    \begin{tikzpicture}[scale=0.45]
         \begin{axis}[
            xlabel={\(\alpha\)},
            ylabel={Accuracy (\%)}, 
            xmin=0.0, xmax=1.0,
            ymin=0, ymax=100,
            xtick={0.0,0.2,0.4,0.6,0.8,1.0},
            ytick={0,20,40,60,80,100},
            ymajorgrids=false,
            legend pos=north east,
            legend style={nodes={scale=0.5, transform shape}},
            cycle list name=color list,
            legend to name=stratdefmtdcompareallsleipnirlegend,
            legend columns=3,
            width=9cm,
            height=8cm
        ]
        
        \addplot
            coordinates {
            (0,87.3)(0.1,84.3)(0.2,82.1)(0.3,78.8)(0.4,76.2)(0.5,73.9)(0.6,71.4)(0.7,69.4)(0.8,66)(0.9,62.4)(1,69.3)
            };
        \addlegendentry{DT}
        \addplot
            coordinates {
            (0,91.5)(0.1,85.7)(0.2,80.8)(0.3,74)(0.4,68.3)(0.5,62.3)(0.6,55.9)(0.7,50.5)(0.8,44)(0.9,37.9)(1,49.9)
            };
        \addlegendentry{RF}
         \addplot
            coordinates {
             (0,91.4)(0.1,83.3)(0.2,76.1)(0.3,67.7)(0.4,59.7)(0.5,50.6)(0.6,43.2)(0.7,35.1)(0.8,27)(0.9,18.8)(1,28.9)
            };
        \addlegendentry{NN-DD}
         \addplot
            coordinates {
           (0,90.1)(0.1,82.7)(0.2,76.5)(0.3,68.4)(0.4,61.2)(0.5,52.8)(0.6,45.6)(0.7,38.4)(0.8,30.9)(0.9,23.3)(1,38.2)
            };
        \addlegendentry{SecSVM}
         \addplot
            coordinates {
            (0,87.7)(0.1,84.8)(0.2,82.6)(0.3,79.3)(0.4,76.3)(0.5,74.3)(0.6,71.6)(0.7,69.4)(0.8,66.2)(0.9,62.7)(1,70.4)
            };
        \addlegendentry{DT-RFN}
         \addplot
            coordinates {
            (0,87.6)(0.1,88.2)(0.2,90.3)(0.3,90.6)(0.4,92.1)(0.5,92.8)(0.6,93.7)(0.7,94.8)(0.8,95.5)(0.9,96.7)(1,95.4)
            };
        \addlegendentry{DT-AT-0.1}
       
         \addplot
            coordinates {
            (0,91.8)(0.1,92.6)(0.2,93.9)(0.3,94.6)(0.4,95.5)(0.5,96)(0.6,96.9)(0.7,97.6)(0.8,98.6)(0.9,99.2)(1,100)
            };
        \addlegendentry{NN-AT-0.25}
         \addplot
            coordinates {
            (0,91.9)(0.1,92.9)(0.2,94.1)(0.3,94.8)(0.4,95.6)(0.5,96.2)(0.6,96.9)(0.7,97.6)(0.8,98.7)(0.9,99.3)(1,100)
            };
        \addlegendentry{RF-AT-0.25}
         \addplot
            coordinates {
            (0,91.9)(0.1,92.9)(0.2,94.1)(0.3,94.8)(0.4,95.6)(0.5,96.2)(0.6,96.9)(0.7,97.6)(0.8,98.7)(0.9,99.3)(1,100)
            };
        \addlegendentry{StratDef-Variety-GT}
         \addplot
            coordinates {
            (0,91.9)(0.1,91.9)(0.2,93.6)(0.3,94)(0.4,95.2)(0.5,95.8)(0.6,96.6)(0.7,97.2)(0.8,98.4)(0.9,99.2)(1,99.5)
            };
        \addlegendentry{StratDef-Variety-Ranked}
        \addplot
            coordinates {
            (0,91.8)(0.1,91.8)(0.2,91.8)(0.3,90.8)(0.4,90.6)(0.5,90.3)(0.6,90.3)(0.7,89.5)(0.8,89.2)(0.9,88.4)(1,91.5)
            };
        \addlegendentry{Voting-Best-Majority}
        
         \addplot
            coordinates {
            (0,91.2)(0.1,92.2)(0.2,93.5)(0.3,94.3)(0.4,95.4)(0.5,95.8)(0.6,96.7)(0.7,97.4)(0.8,98.5)(0.9,99.2)(1,99.9)
            };
        \addlegendentry{Voting-Variety-Majority}
            
        \end{axis}
    \end{tikzpicture}
\caption{SLEIPNIR}
\end{subfigure}

\ref{stratdefmtdcompareallsleipnirlegend}
\caption{Comparison of StratDef with other best-performing defenses against the strong attacker. For SLEIPNIR, the StratDef configurations, NN-AT-0.25 and RF-AT-0.25 have very similar performance.}
\label{figure:mtdcompareall}
\end{figure}
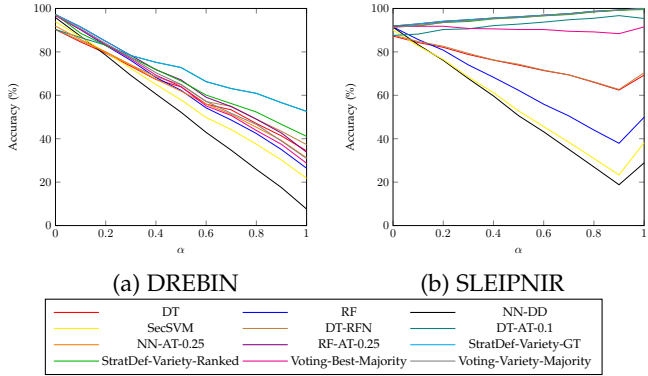

We next compare StratDef with a range of other models and defenses. We present a selection of the results in Figure~\ref{figure:mtdcompareall} covering a number of defenses that we evaluate. In our discussion, we focus on the aggregate results specifically under the strong attacker scenario. For extended results, we refer the reader to Appendix~\ref{appendix:extendedresults}.

Overall, our evaluation shows that the other models and defenses that we evaluate perform significantly worse than StratDef, except for high levels of adversarial training (which we discuss further later). This is especially clear at the highest threat level, with the highest attack intensity and the strongest attacker. For example, as one might expect, the vanilla NN and SVM models perform poorly and only achieve 7\% and 2\% accuracy, respectively, at the peak threat level for DREBIN, with only marginally better performance for SLEIPNIR. However, even after some robustness procedures are applied to these models, the performance only improves by \(\approx\) 10-15\% in accuracy for defensive distillation (NN-DD) and SecSVM. The random feature nullification (RFN) approach offers an insignificant improvement to the vanilla models from all model families, especially, at the peak threat level. These results demonstrate the general ineffectiveness of these defensive approaches. Interestingly, in some instances, the vanilla random forest (RF) for DREBIN and the decision tree (DT) for SLEIPNIR can even surpass defenses such as random feature nullification (RFN), though for SLEIPNIR at the cost of a higher number of false positives. %
Nonetheless, all of these defenses and models still significantly underperform in comparison to our StratDef approach as shown in Figure~\ref{figure:mtdcompareall}, whose average accuracy across the attack intensities sits comfortably above 70\% for both datasets.

StratDef also outperforms the voting defense. At the peak adversarial threat, the Voting-Best-Majority and Voting-Variety-Majority configurations are on par with the vanilla models and other defenses for DREBIN, only achieving a maximum of 30\% accuracy. For SLEIPNIR, although the voting defense can achieve 90+\% accuracy with adequate F1, this comes at the cost of slightly higher FPR and FNR as well as a lower AUC compared with the StratDef Best and Variety configurations (see Appendix~\ref{appendix:extendedresults}). %

Generally, defenses for SLEIPNIR are more consistent and appear to work better because of fewer allowed perturbations, reducing the attack surface and limiting the avenues for evasion. Hence, stronger defenses have it easier to defend themselves against the weaker adversarial examples. Meanwhile, in the more complex scenario with DREBIN, the attacker has greater opportunity with the perturbations as the attacker can perform feature addition and removal, which leads to greater evasion. There, our evaluation shows that StratDef is much superior in dealing with adversarial examples across the attacker scenarios. Intriguingly, from the other defenses we evaluate, we find that only high levels of adversarial training provide any notable robustness. This is similar to findings in previous work \cite{pods, tramer2017ensemble, sengupta2018mtdeep}. For example, at the peak threat level, 50+\% accuracy for DREBIN and 90+\% accuracy for SLEIPNIR can be achieved with high levels of adversarial training across the model families (typically the 0.1-0.25 variants). As a side-effect, however, the adversarially-trained models can sometimes result in more false positives. For example, the highly adversarially-trained models based on the NN and SVM model families for DREBIN, and DT for SLEIPNIR cause more false positive predictions, though this is less observable with the adversarially-trained RF model in both datasets. 

Overall, we observe that highly adversarially-trained RF and DT models are the most balanced all-rounders from existing defenses for both datasets, offering decent F1 and AUC while maintaining fewer false predictions than other defenses. In some cases, while StratDef may perform similarly to such adversarially-trained models, a significant advantage of StratDef is that it simplifies the process of selecting an appropriate model to deploy. Additionally, StratDef has a benefit over single adversarially-trained models as it complicates the attacker's efforts to construct substitute models, reducing the success of black-box attacks as it does not behave like a fixed, static target (see Section~\ref{sec:blackboxattack}).

\subsection{Efficiency of StratDef}
\label{sec:expefficieny}
We also assess the efficiency of StratDef, voting, and some of the other best-performing defenses. This is important, as we hypothesize that defenses may differ in their prediction times and resource consumption due to their particular configuration. For this experiment, we query each considered defense 1000+ times continuously and then measure the time taken to produce a prediction. 

\begin{figure}[!htbp]
\centering
\begin{subfigure}[b]{0.25\textwidth}
    \begin{tikzpicture}[scale=0.45]
        \begin{axis}[
            xlabel={\(\alpha\)},
            ylabel={Average time per prediction (seconds)}, 
            xmin=0.0, xmax=1.0,
            ymin=0,
            xtick={0.0,0.2,0.4,0.6,0.8,1.0},
            ymajorgrids=false,
            legend pos=south east,
            legend style={nodes={scale=0.7, transform shape}},
            cycle list name=color list,
            legend to name=timetakenlegend2,
            legend columns=3,
            ymode=log,
            width=9cm,
            height=8cm
        ]
      
        \addplot
            coordinates {
           (0,0.0005)(0.1,0.0006)(0.2,0.0006)(0.3,0.0006)(0.4,0.0006)(0.5,0.0006)(0.6,0.0006)(0.7,0.0006)(0.8,0.0006)(0.9,0.0006)(1,0.0006)
            };
        \addlegendentry{DT-AT-0.1}

        \addplot
            coordinates {
           (0,0.0083)(0.1,0.0082)(0.2,0.0088)(0.3,0.0091)(0.4,0.0095)(0.5,0.0095)(0.6,0.0101)(0.7,0.0097)(0.8,0.0099)(0.9,0.0105)(1,0.0105)
            };
        \addlegendentry{NN-AT-0.25}
        \addplot
            coordinates {
           (0,0.0213)(0.1,0.0296)(0.2,0.0299)(0.3,0.0303)(0.4,0.0306)(0.5,0.0321)(0.6,0.0305)(0.7,0.0298)(0.8,0.0305)(0.9,0.0312)(1,0.0302)
            };
        \addlegendentry{RF-AT-0.25}
        \addplot
            coordinates {
            (0,0.0213)(0.1,0.017)(0.2,0.0156)(0.3,0.0174)(0.4,0.0166)(0.5,0.019)(0.6,0.0198)(0.7,0.02)(0.8,0.018)(0.9,0.0212)(1,0.0165)
            };
        \addlegendentry{StratDef-Variety-GT}
        \addplot
            coordinates {
           (0,0.062)(0.1,0.061)(0.2,0.0607)(0.3,0.0639)(0.4,0.0644)(0.5,0.0698)(0.6,0.0645)(0.7,0.0626)(0.8,0.0633)(0.9,0.0658)(1,0.0643)
            };
        \addlegendentry{Voting-Variety-Majority}

        \end{axis}
    \end{tikzpicture}
\caption{DREBIN}
\end{subfigure}
\hskip -1ex
\begin{subfigure}[b]{0.22\textwidth}
 \begin{tikzpicture}[scale=0.45]
        \begin{axis}[
            xlabel={\(\alpha\)},
            ylabel={Average time per prediction (seconds)}, 
            xmin=0.0, xmax=1.0,
            ymin=0,
            xtick={0.0,0.2,0.4,0.6,0.8,1.0},
            ymajorgrids=false,
            legend pos=south east,
            legend style={nodes={scale=0.7, transform shape}},
            cycle list name=color list,
            legend to name=timetakenlegend2,
            legend columns=3,
            ymode=log,
            width=9cm,
            height=8cm
        ]
          
        \addplot
            coordinates {
           (0,0.0002)(0.1,0.0004)(0.2,0.0004)(0.3,0.0005)(0.4,0.0005)(0.5,0.0005)(0.6,0.0005)(0.7,0.0005)(0.8,0.0005)(0.9,0.0004)(1,0.0004)
            };
        \addlegendentry{DT-AT-0.1}
   
        \addplot
            coordinates {
           (0,0.004)(0.1,0.0042)(0.2,0.0043)(0.3,0.0047)(0.4,0.0051)(0.5,0.0058)(0.6,0.0057)(0.7,0.0058)(0.8,0.0057)(0.9,0.0055)(1,0.0057)
            };
        \addlegendentry{NN-AT-0.25}
        \addplot
            coordinates {
           (0,0.0181)(0.1,0.0204)(0.2,0.02)(0.3,0.0197)(0.4,0.0202)(0.5,0.0199)(0.6,0.0202)(0.7,0.0205)(0.8,0.0207)(0.9,0.0203)(1,0.0184)
            };
        \addlegendentry{RF-AT-0.25}
        \addplot
            coordinates {
           (0,0.0184)(0.1,0.0135)(0.2,0.0127)(0.3,0.0117)(0.4,0.0114)(0.5,0.0116)(0.6,0.014)(0.7,0.0128)(0.8,0.013)(0.9,0.0137)(1,0.0123)
            };
        \addlegendentry{StratDef-Variety-GT}
        \addplot
            coordinates {
           (0,0.169)(0.1,0.1697)(0.2,0.1696)(0.3,0.1693)(0.4,0.1687)(0.5,0.169)(0.6,0.1689)(0.7,0.1682)(0.8,0.1687)(0.9,0.1682)(1,0.1685)
            };
        \addlegendentry{Voting-Variety-Majority}
        
        \end{axis}
    \end{tikzpicture}
\caption{SLEIPNIR}
\end{subfigure}
\ref{timetakenlegend2}
\caption{Average time taken per prediction against the strong attacker.}
\label{figure:timeresults}
\end{figure}
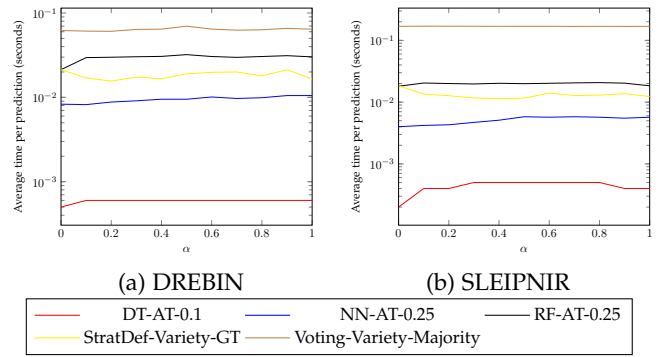

Figure~\ref{figure:timeresults} shows the average time taken by each defense to produce a single prediction against the strong attacker. %
Evidently, there is a significant time difference between StratDef and voting, with voting exhibiting higher costs. This is because, with voting, all constituent models must be waited for while they produce their individual predictions before the final prediction can be returned. Meanwhile, StratDef returns predictions in a similar time to single-model defenses, as only a single constituent model is used for a prediction, with minimal overhead involved in rolling a biased die to choose that particular model at prediction-time. %
In fact, StratDef returns faster predictions on average than RF-AT-0.25 for both datasets. Intriguingly, some prior work has found that random forests are generally slower than other models to produce predictions \cite{bucilua2006model, 8594826, tang2018random}, and in our work, this also seems to be the case in this experimental setting when comparing random forests with StratDef.

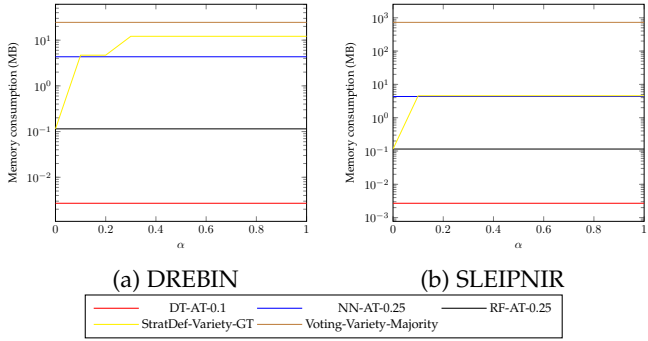
\begin{figure}[!htbp]
\centering
\begin{subfigure}[b]{0.25\textwidth}
    \begin{tikzpicture}[scale=0.45]
        \begin{axis}[
            xlabel={\(\alpha\)},
            ylabel={Memory consumption (MB)}, 
            xmin=0.0, xmax=1.0,
            ymin=0,
            xtick={0.0,0.2,0.4,0.6,0.8,1.0},
            ymajorgrids=false,
            legend pos=south west,
            legend style={nodes={scale=0.5, transform shape}},
            cycle list name=color list,
            legend to name=memorycomparelegend,
            legend columns=3,
            ymode=log,
            width=9cm,
            height=8cm
        ]
        
        \addplot
            coordinates {
           (0,0.0027)(0.1,0.0027)(0.2,0.0027)(0.3,0.0027)(0.4,0.0027)(0.5,0.0027)(0.6,0.0027)(0.7,0.0027)(0.8,0.0027)(0.9,0.0027)(1,0.0027)
            };
        \addlegendentry{DT-AT-0.1}

         \addplot
            coordinates {
           (0,4.3318)(0.1,4.3318)(0.2,4.3318)(0.3,4.3318)(0.4,4.3318)(0.5,4.3318)(0.6,4.3318)(0.7,4.3318)(0.8,4.3318)(0.9,4.3318)(1,4.3318)
           
            };
        \addlegendentry{NN-AT-0.25}
         \addplot
            coordinates {
           (0,0.1146)(0.1,0.1146)(0.2,0.1146)(0.3,0.1146)(0.4,0.1146)(0.5,0.1146)(0.6,0.1146)(0.7,0.1146)(0.8,0.1146)(0.9,0.1146)(1,0.1146)
            };
        \addlegendentry{RF-AT-0.25}
        \addplot
            coordinates {
            (0,0.1146)(0.1,4.6944)(0.2,4.6944)(0.3,12.1498)(0.4,12.1498)(0.5,12.1498)(0.6,12.1498)(0.7,12.1498)(0.8,12.1498)(0.9,12.1498)(1,12.1498)
            };
        \addlegendentry{StratDef-Variety-GT}
        \addplot
            coordinates {
           (0,24.5256)(0.1,24.5256)(0.2,24.5256)(0.3,24.5256)(0.4,24.5256)(0.5,24.5256)(0.6,24.5256)(0.7,24.5256)(0.8,24.5256)(0.9,24.5256)(1,24.5256)
            };
        \addlegendentry{Voting-Variety-Majority}
        
        \end{axis}
    \end{tikzpicture}
\caption{DREBIN}
\end{subfigure}
\hskip -1ex
\begin{subfigure}[b]{0.22\textwidth}
 \begin{tikzpicture}[scale=0.45]
        \begin{axis}[
            xlabel={\(\alpha\)},
            ylabel={Memory consumption (MB)}, 
            xmin=0.0, xmax=1.0,
            ymin=0,
            xtick={0.0,0.2,0.4,0.6,0.8,1.0},
            ymajorgrids=false,
            legend pos=south west,
            legend style={nodes={scale=0.5, transform shape}},
            cycle list name=color list,
            legend to name=memorycomparelegend,
            legend columns=3,
            ymode=log,
            width=9cm,
            height=8cm
        ]
        
        \addplot
            coordinates {
           (0,0.0027)(0.1,0.0027)(0.2,0.0027)(0.3,0.0027)(0.4,0.0027)(0.5,0.0027)(0.6,0.0027)(0.7,0.0027)(0.8,0.0027)(0.9,0.0027)(1,0.0027)
            };
        \addlegendentry{DT-AT-0.1}

         \addplot
            coordinates {
           (0,4.3322)(0.1,4.3322)(0.2,4.3322)(0.3,4.3322)(0.4,4.3322)(0.5,4.3322)(0.6,4.3322)(0.7,4.3322)(0.8,4.3322)(0.9,4.3322)(1,4.3322)
            };
        \addlegendentry{NN-AT-0.25}
           \addplot
            coordinates {
           (0,0.1146)(0.1,0.1146)(0.2,0.1146)(0.3,0.1146)(0.4,0.1146)(0.5,0.1146)(0.6,0.1146)(0.7,0.1146)(0.8,0.1146)(0.9,0.1146)(1,0.1146)
            };
        \addlegendentry{RF-AT-0.25}
        \addplot
            coordinates {
            (0,0.1146)(0.1,4.5499)(0.2,4.5499)(0.3,4.5499)(0.4,4.5499)(0.5,4.5499)(0.6,4.5499)(0.7,4.5499)(0.8,4.5499)(0.9,4.5499)(1,4.5499)
            };
        \addlegendentry{StratDef-Variety-GT}
        \addplot
            coordinates {
           (0,728.3915)(0.1,728.3915)(0.2,728.3915)(0.3,728.3915)(0.4,728.3915)(0.5,728.3915)(0.6,728.3915)(0.7,728.3915)(0.8,728.3915)(0.9,728.3915)(1,728.3915)
            };
        \addlegendentry{Voting-Variety-Majority}
      
        \end{axis}
    \end{tikzpicture}
\caption{SLEIPNIR}
\end{subfigure}
\ref{memorycomparelegend}
\caption{Average memory consumption of models against the strong attacker.}
\label{figure:memoryresults}
\end{figure}

We also examine the average memory consumption incurred when producing a prediction against the strong attacker. This is because we are interested in analyzing the overhead of keeping models in memory to serve predictions. Figure~\ref{figure:memoryresults} shows the average memory consumption of each defense across attack intensities. Defenses besides StratDef have static memory consumption because they are not strategized for the attack intensity. The single-model defenses we evaluate consume less memory, with minor differences due to the particular model family. Meanwhile, ensemble defenses require access to more models at prediction-time, leading to higher memory costs. However, voting has the highest memory consumption since it uses all models in the ensemble for a single prediction. For example, in the case of Voting-Variety-Majority for SLEIPNIR, a single, memory-intensive model requires over 0.7GB memory. Meanwhile, StratDef is efficient and better than voting, as it only loads models for each attack intensity with a non-zero probability (i.e., those that have a non-zero chance of being chosen to make a prediction) rather than keeping all constituent models in memory. Overall, StratDef --- which is an ensemble defense --- performs as efficiently as (and sometimes better than) single-model defenses, considering both time and memory costs. 

\subsection{StratDef vs. UAPs}
\label{sec:dealingwithuaps}
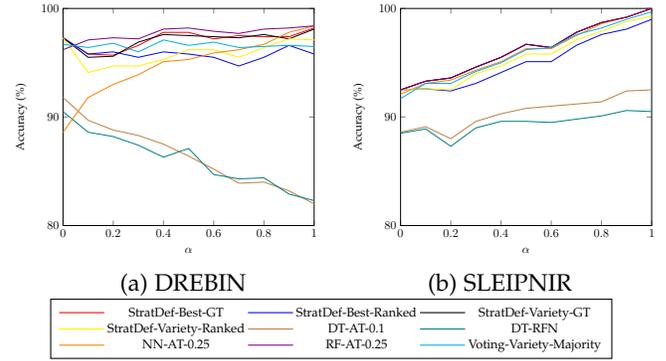
\begin{figure}[!htbp]
\centering
\begin{subfigure}[b]{0.25\textwidth}
    \begin{tikzpicture}[scale=0.45]
        \begin{axis}[
            xlabel={\(\alpha\)},
            ylabel={Accuracy (\%)}, 
            xmin=0.0, xmax=1.0,
            ymin=80, ymax=100,
            xtick={0.0,0.2,0.4,0.6,0.8,1.0},
            ytick={80,90,100},
            ymajorgrids=false,
            legend pos=south west,
            legend style={nodes={scale=0.5, transform shape}},
            cycle list name=color list,
            legend to name=mtduapslegend,
            legend columns=3,
            width=9cm,
            height=8cm
        ]

         \addplot
            coordinates {
             (0,97.3)(0.1,95.8)(0.2,95.7)(0.3,96.6)(0.4,97.8)(0.5,97.8)(0.6,97.2)(0.7,97.5)(0.8,97.4)(0.9,97.4)(1,98.2)
            };
        \addlegendentry{StratDef-Best-GT}
       
         \addplot
            coordinates {
             (0,97.3)(0.1,95.8)(0.2,96)(0.3,95.5)(0.4,96)(0.5,95.8)(0.6,95.5)(0.7,94.7)(0.8,95.5)(0.9,96.6)(1,95.8)
            };
        \addlegendentry{StratDef-Best-Ranked}
        
         \addplot
            coordinates {
             (0,97.3)(0.1,95.5)(0.2,95.6)(0.3,96.9)(0.4,97.6)(0.5,97.5)(0.6,97.4)(0.7,97.3)(0.8,97.6)(0.9,97.2)(1,98.1)
            };
        \addlegendentry{StratDef-Variety-GT}

         \addplot
            coordinates {
             (0,97.3)(0.1,94.1)(0.2,94.7)(0.3,94.7)(0.4,95.3)(0.5,96.2)(0.6,96.2)(0.7,95.5)(0.8,96.4)(0.9,97.2)(1,97.1)
            };
        \addlegendentry{StratDef-Variety-Ranked}
        
         \addplot
            coordinates {
             (0,91.8)(0.1,89.7)(0.2,88.8)(0.3,88.3)(0.4,87.5)(0.5,86.4)(0.6,85.2)(0.7,83.9)(0.8,84)(0.9,83.2)(1,82)
            };
        \addlegendentry{DT-AT-0.1}
        
         \addplot
            coordinates {
             (0,90.5)(0.1,88.6)(0.2,88.2)(0.3,87.4)(0.4,86.3)(0.5,87.1)(0.6,84.7)(0.7,84.3)(0.8,84.4)(0.9,82.9)(1,82.3)
            };
        \addlegendentry{DT-RFN}

         \addplot
            coordinates {
             (0,88.6)(0.1,91.8)(0.2,93)(0.3,93.9)(0.4,95.1)(0.5,95.3)(0.6,95.9)(0.7,96.2)(0.8,96.7)(0.9,97.8)(1,98.4)
            };
        \addlegendentry{NN-AT-0.25}
        
         \addplot
            coordinates {
             (0,96.2)(0.1,97.1)(0.2,97.3)(0.3,97.2)(0.4,98.1)(0.5,98.2)(0.6,97.9)(0.7,97.7)(0.8,98.1)(0.9,98.2)(1,98.4)
            };
        \addlegendentry{RF-AT-0.25}
        
        \addplot
            coordinates {
             (0,96.7)(0.1,96.4)(0.2,96.8)(0.3,96)(0.4,97.1)(0.5,96.6)(0.6,96.9)(0.7,96.4)(0.8,96.5)(0.9,96.6)(1,96.5)
            };
        \addlegendentry{Voting-Variety-Majority}

        \end{axis}
    \end{tikzpicture}
\caption{DREBIN}
\label{figure:mtduapspt1}
\end{subfigure}
\hskip -1ex
\begin{subfigure}[b]{0.22\textwidth}
    \begin{tikzpicture}[scale=0.45]
        \begin{axis}[
            xlabel={\(\alpha\)},
            ylabel={Accuracy (\%)}, 
            xmin=0.0, xmax=1.0,
            ymin=80, ymax=100,
            xtick={0.0,0.2,0.4,0.6,0.8,1.0},
            ytick={80,90,100},
            ymajorgrids=false,
            legend pos=south west,
            legend style={nodes={scale=0.5, transform shape}},
            cycle list name=color list,
            legend to name=mtduapslegend,
            legend columns=3,
            width=9cm,
            height=8cm
        ]
     
         \addplot
            coordinates {
             (0,92.5)(0.1,93.3)(0.2,93.6)(0.3,94.6)(0.4,95.5)(0.5,96.7)(0.6,96.4)(0.7,97.8)(0.8,98.7)(0.9,99.2)(1,100)
            };
        \addlegendentry{StratDef-Best-GT}
        
         \addplot
            coordinates {
             (0,92.5)(0.1,92.6)(0.2,92.4)(0.3,93.1)(0.4,94.1)(0.5,95.1)(0.6,95.1)(0.7,96.6)(0.8,97.6)(0.9,98.1)(1,99)
            };
        \addlegendentry{StratDef-Best-Ranked}

         \addplot
            coordinates {
             (0,92.5)(0.1,93.3)(0.2,93.6)(0.3,94.6)(0.4,95.5)(0.5,96.7)(0.6,96.4)(0.7,97.8)(0.8,98.7)(0.9,99.2)(1,100)
            };
        \addlegendentry{StratDef-Variety-GT}

         \addplot
            coordinates {
             (0,92.5)(0.1,92.6)(0.2,92.5)(0.3,94)(0.4,94.7)(0.5,95.8)(0.6,95.8)(0.7,97.1)(0.8,97.9)(0.9,98.8)(1,99.3)
            };
        \addlegendentry{StratDef-Variety-Ranked}
        
         \addplot
            coordinates {
                (0,88.6)(0.1,89.1)(0.2,88)(0.3,89.6)(0.4,90.3)(0.5,90.8)(0.6,91)(0.7,91.2)(0.8,91.4)(0.9,92.4)(1,92.5)
            };
        \addlegendentry{DT-AT-0.1}
        \addplot
            coordinates {
            (0,88.5)(0.1,88.9)(0.2,87.3)(0.3,89)(0.4,89.6)(0.5,89.6)(0.6,89.5)(0.7,89.8)(0.8,90.1)(0.9,90.6)(1,90.5)
            };
        \addlegendentry{DT-RFN}

         \addplot
            coordinates {
            (0,92.1)(0.1,93.1)(0.2,93.4)(0.3,94.3)(0.4,95.1)(0.5,96.3)(0.6,96.3)(0.7,97.5)(0.8,98.6)(0.9,99.2)(1,100)
            };
        \addlegendentry{NN-AT-0.25}
        
         \addplot
            coordinates {
                (0,92.5)(0.1,93.3)(0.2,93.6)(0.3,94.6)(0.4,95.5)(0.5,96.7)(0.6,96.4)(0.7,97.8)(0.8,98.7)(0.9,99.2)(1,100)
            };
        \addlegendentry{RF-AT-0.25}
        
        \addplot
            coordinates {
                (0,91.7)(0.1,93.1)(0.2,93.1)(0.3,94.2)(0.4,95)(0.5,96.2)(0.6,96.4)(0.7,97.6)(0.8,98.2)(0.9,99)(1,99.7)
            };
        \addlegendentry{Voting-Variety-Majority}
            
        \end{axis}
    \end{tikzpicture}
\caption{SLEIPNIR}
\label{figure:mtduapspt2}
\end{subfigure}
\ref{mtduapslegend}
\caption{Accuracy of different defenses against universal attacker. For SLEIPNIR, StratDef-Best-GT and StratDef-Variety-GT have same performance. Some models have similar performance --- see Appendix~\ref{appendix:extendedresults} for extended results.}
\label{figure:mtduaps}

\end{figure}

Recent work has uncovered universal adversarial perturbations (UAPs) as a cost-effective method of generating adversarial examples where perturbations can be reused \cite{moosavi2017universal, labaca2021universal}. We therefore evaluate how StratDef performs against the \emph{universal attacker}, who only uses adversarial examples generated from UAPs. For this, we determine if a set of perturbations has been precisely reused to generate adversarial examples (from Section~\ref{sec:crafting}). In such situations, adversarial examples are regarded as having been generated by UAPs. For example, if the application of the same set of perturbations to several distinct malware samples has produced adversarial examples, the set of perturbations counts as UAPs. In total, there are 1,541 such adversarial examples for DREBIN (\(UAP_{DREBIN}\)) and 2,217 for SLEIPNIR (\(UAP_{SLEIPNIR}\)). We present the results of this experiment in Figure~\ref{figure:mtduaps} (with Appendix~\ref{appendix:extendedresults} containing all results).

Figure~\ref{figure:mtduaps} shows the accuracy of the StratDef configurations and other best-performing models against the universal attacker. The adversarial examples in \(UAP_{DREBIN}\) and \(UAP_{SLEIPNIR}\) appear less evasive than before, as the accuracy of most models is hardly affected. That is, the universal attacker is unable to achieve effective evasion against most defenses, including any StratDef configuration, with its accuracy remaining largely over 90\%. Similar to the previous sections, we observe that the performance of some defenses increases as $\alpha$ increases. This is due to the increased number of weaker adversarial examples that are detected more easily by the defenses. Only vanilla and weakly-defended NN and SVM models fall prey to the universal attacker, with their accuracy reduced to less than \(\approx\) 40\% across both datasets (these are not shown in Figure~\ref{figure:mtduaps} but are available in Appendix~\ref{appendix:extendedresults}). Evidently, the results are more closely concentrated for StratDef, with generally sound performance across both datasets, %
particularly in its GT configurations.

\subsection{StratDef vs. Black-box Attacks}%
\label{sec:blackboxattack}
We also explore how StratDef performs against a complete black-box %
attack. In this setting, a zero-knowledge black-box attacker queries the target model (e.g., StratDef) as an oracle to develop a substitute model \cite{papernot2017practical}. The substitute model is then attacked in anticipation that adversarial examples transfer to the target model.  Hence, the adversarial examples generated in this attack are of an unknown nature; for example, StratDef is not strategized to deal with the attack, nor are any models adversarially-trained on them.

\noindent{\textbf{Procedure.}} For this experiment, we follow the standard procedure to attack a target model using a black-box transferability attack strategy. This is based on the well-established idea that adversarial examples generated for a substitute model may transfer to the target model \cite{szegedy2013intriguing}. 

\begin{figure}[!htbp]
\centering
\begin{subfigure}[b]{0.25\textwidth}
    \begin{tikzpicture}[scale=0.45]
        \begin{axis}[
            xlabel={Number of samples at training-time},
            ylabel={Evasion rate (\%)}, 
            xmin=2, xmax=500,
            ymin=0, ymax=30,
            xtick={2,100,200,300,400,500},
            ytick={0,10,20,30},
            ymajorgrids=false,
            legend pos=south east,
            legend style={nodes={scale=0.5, transform shape}},
            cycle list name=color list,
            legend to name=blackboxlegend,
            legend columns=3,
            width=9cm,
            height=8cm
        ]
        \addplot
            coordinates {
				(2,9.279)(100,19.105)(200,19.105)(300,19.214)(400,19.214)(500,18.886)
            };
        \addlegendentry{DT-AT-0.1}
        \addplot
            coordinates {
				(2,8.843)(100,23.69)(200,23.799)(300,24.017)(400,24.236)(500,24.454)
            };
        \addlegendentry{NN-AT-0.25}
        \addplot
            coordinates {
				(2,5.349)(100,21.834)(200,21.834)(300,22.817)(400,23.472)(500,23.69)
            };
        \addlegendentry{RF-AT-0.25}
        \addplot
            coordinates {
				(2,6.004)(100,18.777)(200,18.231)(300,17.031)(400,18.777)(500,18.996)
            };
        \addlegendentry{StratDef-Variety-GT}
        \addplot
            coordinates {
				(2,9.279)(100,22.598)(200,22.598)(300,22.489)(400,22.926)(500,23.144)
            };
        \addlegendentry{Voting-Variety-Majority}
        
        \end{axis}
    \end{tikzpicture}
\caption{DREBIN}
\end{subfigure}
\hskip -1ex
\begin{subfigure}[b]{0.22\textwidth}
 \begin{tikzpicture}[scale=0.45]
        \begin{axis}[
            xlabel={Number of samples at training-time},
            ylabel={Evasion rate (\%)}, 
            xmin=2, xmax=500,
            ymin=0, ymax=30,
            xtick={2,100,200,300,400,500},
            ytick={0,10,20,30},
            ymajorgrids=false,
            legend pos=south east,
            legend style={nodes={scale=0.5, transform shape}},
            cycle list name=color list,
            legend to name=blackboxlegend,
            legend columns=3,
            width=9cm,
            height=8cm
        ]
        
        \addplot
            coordinates {
				(2,2.826)(100,27.174)(200,26.739)(300,28.043)(400,26.848)(500,25.87)
            };
        \addlegendentry{DT-AT-0.1}
        \addplot
            coordinates {
				(2,0.652)(100,0.109)(200,0.109)(300,0.109)(400,0.109)(500,0.435)
            };
        \addlegendentry{NN-AT-0.25}
        \addplot
            coordinates {
				(2,1.196)(100,0)(200,0)(300,0)(400,0)(500,0)
            };
        \addlegendentry{RF-AT-0.25}
        \addplot
            coordinates {
				(2,1.087)(100,2.174)(200,2.826)(300,0.217)(400,2.5)(500,0.978)
            };
        \addlegendentry{StratDef-Variety-GT}
        \addplot
            coordinates {
				(2,0.978)(100,3.913)(200,3.804)(300,0.652)(400,0.326)(500,0.435)
            };
        \addlegendentry{Voting-Variety-Majority}
        
        \end{axis}
    \end{tikzpicture}
\caption{SLEIPNIR}
\end{subfigure}
\ref{blackboxlegend}
\caption{Results of black-box attack against various defenses.}
\label{figure:blackboxsucc}
\end{figure}
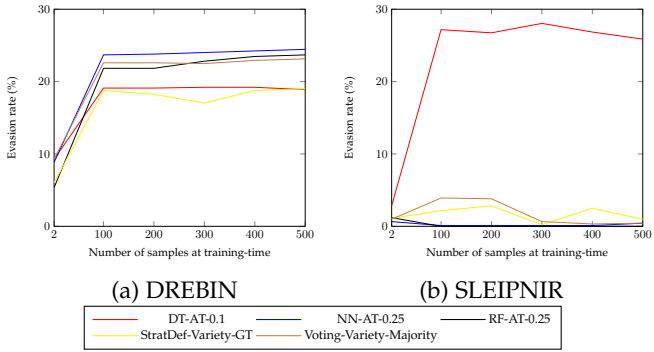

Initially, we query the target model (whether it is StratDef or some other defense) with an equal number of benign and malware samples\footnote{Therefore, Figure~\ref{figure:blackboxsucc} starts with 2 samples because a single sample from each class is used to build the training set for the substitute DNN.} and record the predicted outputs from the model. We vary the \emph{number of [these] samples at training-time} to examine if this affects the success of attacks, as more interactions at training-time should produce a better representation of the target model. The input-output relations from querying the target model are then used to train a substitute DNN (see Appendix~\ref{appendix:architecutres} for the model architecture), which acts as an estimation of the target model. Against this substitute DNN, we then use white-box attacks (BIM \cite{kurakin2016adversarial}, FGSM \cite{goodfellow2014explaining, dhaliwal2018gradient}, JSMA \cite{papernot2016limitations} and PGD \cite{madry2017towards}) to generate adversarial examples, in anticipation that they will transfer to the target model.  As before, procedures are applied to ensure a lower bound of functionality preservation in the feature-space when generating the adversarial examples (see Section~\ref{sec:crafting}). The adversarial examples are then tested against the target models, such as StratDef and other defenses.

In this attack setting, StratDef assumes the highest threat level (i.e., the strongest attacker at the highest attack intensity). Beyond this, StratDef is not strategized to deal with a black-box attack of any kind. Therefore, this attack also helps us understand how StratDef may work against an unknown attacker. %

\noindent{\textbf{Results.}} 
In Figure~\ref{figure:blackboxsucc}, we present the results of this experiment, where we compare StratDef with other well-performing defenses from the prior subsections. We find that StratDef works well across both datasets, with a relatively low evasion rate compared with other defenses. For DREBIN, the attacker achieves a 19\% evasion rate against StratDef in the worst-case which is lower, and hence better, than other defenses, and around 16\% evasion rate on average across all results, which is still lower than the other defenses. Recall that StratDef is not currently strategized to deal with such an attack despite its better performance. As StratDef cycles between models during predictions, we also observe variations in the attacker's performance.  Meanwhile, although the adversarially-trained DT model performs adequately against DREBIN, it performs much worse for SLEIPNIR, while other models also exhibit mixed performance. Regardless, in the more complex scenario involving DREBIN, StratDef offers superior performance against black-box attacks. There, we generally observe that as the number of samples at training-time increases, the evasion rate also increases up to a certain level. This supports the hypothesis that substitute models that are trained using a higher number of input-output relations of the target model are better representations of it.

For SLEIPNIR, the attacker is less successful, which is a theme we have seen previously. This is due to a more limited feature-space  (i.e., the set of allowed perturbations for generating adversarial examples) and is reflected in the results for the black-box attack, where the evasion rate decreases considerably against the stronger defenses such as StratDef (\(< 1\%\) evasion rate). However, the DT-based model is greatly evaded, with an evasion rate of 25+\%. StratDef makes use of the DT-AT-0.1 model in its strategy at the considered threat level and therefore suffers slightly in comparison to other models.

\section{Discussion \& Limitations}
\label{sec:discussion}
In this section, we discuss the overall findings from our evaluation of StratDef and the other defenses. This is followed by a discussion on the limitations we see of our work.

In contrast to prior work, StratDef selects the best models to use in an MTD in a way that is both dynamic and strategic. We focus on the fundamental aspects of the design of MTDs, such as model construction, model selection, and optimizer selection. This helps not only to increase accuracy and the level of uncertainty for attackers, but also reduces the transferability of adversarial examples, as demonstrated by our results. In particular, the StratDef configurations using the Variety model selection perform better than other configurations, as this model selection enforces diversity among the constituent models which offsets transferability better. When this model selection is paired with the game-theoretic (GT) optimizer, StratDef exhibits peak performance against adversarial attacks. Intriguingly, however, the GT configurations sometimes use pure strategies (and therefore use fewer models), which may increase the risk of an attacker profiling the defense. This is in contrast with the strategic ranked optimizer (Ranked), which uses more constituent models but can lead to greater evasion due to transferability, owing to an increased chance that a vulnerable constituent model is chosen at prediction-time. Hence, there is a trade-off between the number of models from the ensemble that are used and the robustness of the system, with the attacker's success reduced only with diverse constituent models. Nonetheless, both of these optimizers are far superior to the uniform random strategy (URS) approach, demonstrating the importance of proper prediction-time strategies irrespective of the soundness of the model selection.

When compared with other models and defenses, our results show that StratDef performs better than the majority of them across the different threat levels. Defenses such as defensive distillation, random feature nullification, and SecSVM exhibit minimal adversarial robustness, with performance that is quite comparable to that of their vanilla counterparts, especially at the peak threat level. Meanwhile, despite using the same model selections, voting defense is less successful against attacks than StratDef and more costly to deploy, especially in the more complex scenario with DREBIN. This demonstrates the superiority of a strategic MTD approach. Intriguingly, from the existing defenses, only high levels of adversarial training provide effective robustness against attacks. While in some instances, StratDef may perform similarly to adversarially-trained models, an advantage is that StratDef significantly simplifies the process of selecting an appropriate model to deploy. Additionally, StratDef complicates the attacker's efforts to construct substitute models, making black-box attacks less successful than against other defenses, as it does not behave like a fixed, static target as other defenses do.

Despite showing promise against adversarial attacks, we see some limitations for our defense. Depending on the optimizer (e.g., the game-theoretic optimizer), a pure strategy may be developed, which is akin to employing a single constituent model. This may reduce the benefits of an MTD approach and increase the risk of the system being profiled. It would be interesting to explore whether the optimizer could be developed to limit the phenomenon. As discussed, however, using mixed strategies with more models can widen the attack surface.

StratDef is not currently strategized to deal with black-box attacks, such as black-box substitute model attacks (e.g., \cite{papernot2017practical}). In an experiment with such an attack, StratDef exhibited greater robustness than most models. However, there is still scope for improvement in its performance against such an attack. For example, methods to limit how an attacker could obtain information for constructing substitute models in the first place could be explored.

Furthermore, while StratDef aims to prevent the evasion of the model itself by an individual adversarial example, it cannot detect whether an attack is in progress. This is especially pertinent in light of the threat posed by \emph{query attacks} (e.g., \cite{rosenberg2020query, chen2020hopskipjumpattack}). Such attacks generate adversarial examples by iteratively perturbing input samples towards the desired class through feedback received from the target model, rather than through substitute models. Against such attacks, \emph{stateful defenses} (e.g., \cite{chen2020stateful}) have been proposed that monitor and analyze the distribution of queries received by the system. It would be interesting to examine how StratDef could be developed or combined with such stateful detection systems to mitigate these attacks.

Additionally, while the focus of our work has been in the ML-based malware detection domain, it would be interesting to examine whether StratDef, or a similar strategic MTD, could produce similar promising results in other domains.

\section{Conclusion}
\label{sec:conclusion}
In this paper, we presented our strategic defense, StratDef, for defending against adversarial attacks for ML-based malware detection. We have demonstrated the superiority of StratDef over existing defenses across both Android and Windows malware. StratDef embraces the key design principles of a moving target defense and provides a complete framework for building a strategic ensemble defense using different heuristically-driven methods for determining what, how and when a moving target defense system should move to achieve a high degree of adversarial robustness. We have illustrated the dynamic nature of StratDef, which offers flexible methods to promote model heterogeneity, adversarial robustness, and accurate predictions. Moreover, we have shown how StratDef can adapt to the threat level based on the information it has about its operating environment. Experimentally, we have demonstrated StratDef's ability to achieve high levels of adversarial robustness across different threat levels without compromising on performance when compared with other defenses. Overall, we have demonstrated the ability to construct a strategic defense that can increase accuracy by 50+\% while reducing the success of targeted attacks by increasing the uncertainty and complexity for the attacker. 

The results in this paper motivate and provide evidence supporting a strategic MTD approach for dealing with adversarial examples in the malware detection domain. Beyond the work presented in this paper, multiple avenues exist for future work on strategic defenses in this domain. For example, %
we plan to investigate how to deal with black-box attacks even better. This may be achieved by adapting the defense strategy according to the current perceived threat levels that could be based on automated, stateful approaches \cite{chen2020stateful}, or on cyber-threat intelligence~\cite{shu2018threat,zhu2018chainsmith}. %

\ifCLASSOPTIONcaptionsoff
  \newpage
\fi

\bibliographystyle{IEEEtran}
\bibliography{bib.bib}

\begin{thebibliography}{100}
\providecommand{\url}[1]{#1}
\csname url@samestyle\endcsname
\providecommand{\newblock}{\relax}
\providecommand{\bibinfo}[2]{#2}
\providecommand{\BIBentrySTDinterwordspacing}{\spaceskip=0pt\relax}
\providecommand{\BIBentryALTinterwordstretchfactor}{4}
\providecommand{\BIBentryALTinterwordspacing}{\spaceskip=\fontdimen2\font plus
\BIBentryALTinterwordstretchfactor\fontdimen3\font minus
  \fontdimen4\font\relax}
\providecommand{\BIBforeignlanguage}[2]{{%
\expandafter\ifx\csname l@#1\endcsname\relax
\typeout{** WARNING: IEEEtran.bst: No hyphenation pattern has been}%
\typeout{** loaded for the language `#1'. Using the pattern for}%
\typeout{** the default language instead.}%
\else
\language=\csname l@#1\endcsname
\fi
#2}}
\providecommand{\BIBdecl}{\relax}
\BIBdecl

\bibitem{he2015delving}
K.~He, X.~Zhang, S.~Ren, and J.~Sun, ``Delving deep into rectifiers: Surpassing
  human-level performance on imagenet classification,'' in \emph{Proceedings of
  the IEEE international conference on computer vision}, 2015, pp. 1026--1034.

\bibitem{chio2018machine}
C.~Chio and D.~Freeman, \emph{Machine learning and security: Protecting systems
  with data and algorithms}.\hskip 1em plus 0.5em minus 0.4em\relax " O'Reilly
  Media, Inc.", 2018.

\bibitem{Eykholt_2018_CVPR}
K.~Eykholt, I.~Evtimov, E.~Fernandes, B.~Li, A.~Rahmati, C.~Xiao, A.~Prakash,
  T.~Kohno, and D.~Song, ``Robust physical-world attacks on deep learning
  visual classification,'' in \emph{Proceedings of the IEEE Conference on
  Computer Vision and Pattern Recognition (CVPR)}, June 2018.

\bibitem{evtimov2017robust}
I.~Evtimov, K.~Eykholt, E.~Fernandes, T.~Kohno, B.~Li, A.~Prakash, A.~Rahmati,
  and D.~Song, ``Robust physical-world attacks on machine learning models,''
  \emph{arXiv preprint arXiv:1707.08945}, vol.~2, no.~3, p.~4, 2017.

\bibitem{chakraborty2018adversarial}
A.~Chakraborty, M.~Alam, V.~Dey, A.~Chattopadhyay, and D.~Mukhopadhyay,
  ``Adversarial attacks and defences: A survey,'' \emph{arXiv preprint
  arXiv:1810.00069}, 2018.

\bibitem{tramer2017ensemble}
F.~Tram{\`e}r, A.~Kurakin, N.~Papernot, I.~Goodfellow, D.~Boneh, and
  P.~McDaniel, ``Ensemble adversarial training: Attacks and defenses,''
  \emph{arXiv preprint arXiv:1705.07204}, 2017.

\bibitem{xu2017feature}
W.~Xu, D.~Evans, and Y.~Qi, ``Feature squeezing: Detecting adversarial examples
  in deep neural networks,'' \emph{arXiv preprint arXiv:1704.01155}, 2017.

\bibitem{wang2016random}
Q.~Wang, W.~Guo, K.~Zhang, X.~Xing, C.~L. Giles, and X.~Liu, ``Random feature
  nullification for adversary resistant deep architecture,'' \emph{arXiv
  preprint arXiv:1610.01239}, 2016.

\bibitem{xie2017mitigating}
C.~Xie, J.~Wang, Z.~Zhang, Z.~Ren, and A.~Yuille, ``Mitigating adversarial
  effects through randomization,'' \emph{arXiv preprint arXiv:1711.01991},
  2017.

\bibitem{sengupta2018mtdeep}
S.~Sengupta, T.~Chakraborti, and S.~Kambhampati, ``Mtdeep: boosting the
  security of deep neural nets against adversarial attacks with moving target
  defense,'' in \emph{Workshops at the Thirty-Second AAAI Conference on
  Artificial Intelligence}, 2018.

\bibitem{chen2018automated}
S.~Chen, M.~Xue, L.~Fan, S.~Hao, L.~Xu, H.~Zhu, and B.~Li, ``Automated
  poisoning attacks and defenses in malware detection systems: An adversarial
  machine learning approach,'' \emph{computers \& security}, vol.~73, pp.
  326--344, 2018.

\bibitem{pierazzi2020problemspace}
\BIBentryALTinterwordspacing
F.~Pierazzi, F.~Pendlebury, J.~Cortellazzi, and L.~Cavallaro, ``Intriguing
  properties of adversarial ml attacks in the problem space,'' in \emph{2020
  IEEE Symposium on Security and Privacy (SP)}.\hskip 1em plus 0.5em minus
  0.4em\relax IEEE Computer Society, 2020, pp. 1308--1325. [Online]. Available:
  \url{https://doi.ieeecomputersociety.org/10.1109/SP40000.2020.00073}
\BIBentrySTDinterwordspacing

\bibitem{carlinilist}
\BIBentryALTinterwordspacing
N.~Carlini. A complete list of all (arxiv) adversarial example papers.
  [Online]. Available:
  \url{https://nicholas.carlini.com/writing/2019/all-adversarial-example-papers.html}
\BIBentrySTDinterwordspacing

\bibitem{pods}
R.~Podschwadt and H.~Takabi, ``{On Effectiveness of Adversarial Examples and
  Defenses for Malware Classification}.'' \emph{International Conference on
  Security and Privacy in Communication Systems}, pp. 380--393, 2019.

\bibitem{szegedy2013intriguing}
C.~Szegedy, W.~Zaremba, I.~Sutskever, J.~Bruna, D.~Erhan, I.~Goodfellow, and
  R.~Fergus, ``Intriguing properties of neural networks,'' \emph{arXiv preprint
  arXiv:1312.6199}, 2014.

\bibitem{papernot2016distillation}
N.~Papernot, P.~McDaniel, X.~Wu, S.~Jha, and A.~Swami, ``Distillation as a
  defense to adversarial perturbations against deep neural networks,'' in
  \emph{2016 IEEE Symposium on Security and Privacy (SP)}.\hskip 1em plus 0.5em
  minus 0.4em\relax IEEE, 2016, pp. 582--597.

\bibitem{mtddef}
\BIBentryALTinterwordspacing
D.~of~Homeland~Security. Moving target defense. [Online]. Available:
  \url{https://www.dhs.gov/science-and-technology/csd-mtd}
\BIBentrySTDinterwordspacing

\bibitem{cho2020toward}
J.-H. Cho, D.~P. Sharma, H.~Alavizadeh, S.~Yoon, N.~Ben-Asher, T.~J. Moore,
  D.~S. Kim, H.~Lim, and F.~F. Nelson, ``Toward proactive, adaptive defense: A
  survey on moving target defense,'' \emph{IEEE Communications Surveys \&
  Tutorials}, vol.~22, no.~1, pp. 709--745, 2020.

\bibitem{qian2020ei}
\BIBentryALTinterwordspacing
Y.~Qian, Y.~Guo, Q.~Shao, J.~Wang, B.~Wang, Z.~Gu, X.~Ling, and C.~Wu,
  ``Ei-mtd: Moving target defense for edge intelligence against adversarial
  attacks,'' \emph{ACM Trans. Priv. Secur.}, vol.~25, no.~3, may 2022.
  [Online]. Available: \url{https://doi.org/10.1145/3517806}
\BIBentrySTDinterwordspacing

\bibitem{amich2021morphence}
\BIBentryALTinterwordspacing
A.~Amich and B.~Eshete, ``Morphence: Moving target defense against adversarial
  examples,'' in \emph{Annual Computer Security Applications Conference}, ser.
  ACSAC.\hskip 1em plus 0.5em minus 0.4em\relax New York, NY, USA: Association
  for Computing Machinery, 2021, p. 61–75. [Online]. Available:
  \url{https://doi.org/10.1145/3485832.3485899}
\BIBentrySTDinterwordspacing

\bibitem{song2019moving}
Q.~Song, Z.~Yan, and R.~Tan, ``Moving target defense for deep visual sensing
  against adversarial examples,'' \emph{arXiv preprint arXiv:1905.13148}, 2019.

\bibitem{9833895}
F.~Ahmed, P.~Vaishnavi, K.~Eykholt, and A.~Rahmati, ``Ares: A system-oriented
  wargame framework for adversarial ml,'' in \emph{2022 IEEE Security and
  Privacy Workshops (SPW)}, 2022, pp. 73--79.

\bibitem{additional1}
\BIBentryALTinterwordspacing
A.~Roy, A.~Chhabra, C.~A. Kamhoua, and P.~Mohapatra, ``A moving target defense
  against adversarial machine learning,'' in \emph{Proceedings of the 4th
  ACM/IEEE Symposium on Edge Computing}, ser. SEC '19.\hskip 1em plus 0.5em
  minus 0.4em\relax New York, NY, USA: Association for Computing Machinery,
  2019, p. 383–388. [Online]. Available:
  \url{https://doi.org/10.1145/3318216.3363338}
\BIBentrySTDinterwordspacing

\bibitem{additional2}
\BIBentryALTinterwordspacing
R.~Izmailov, P.~Lin, S.~Venkatesan, and S.~Sugrim, \emph{Combinatorial Boosting
  of Classifiers for Moving Target Defense Against Adversarial Evasion
  Attacks}.\hskip 1em plus 0.5em minus 0.4em\relax New York, NY, USA:
  Association for Computing Machinery, 2021, p. 13–21. [Online]. Available:
  \url{https://doi.org/10.1145/3474370.3485661}
\BIBentrySTDinterwordspacing

\bibitem{wang2020mtdnnf}
W.~Wang, X.~Xiong, S.~Wang, and J.~Zhang, ``Mtdnnf: Building the security
  framework for deep neural network by moving target defense,'' in \emph{2020
  3rd International Conference on Algorithms, Computing and Artificial
  Intelligence}, 2020, pp. 1--1.

\bibitem{shahzad2013comparative}
R.~K. Shahzad and N.~Lavesson, ``Comparative analysis of voting schemes for
  ensemble-based malware detection,'' \emph{Journal of Wireless Mobile
  Networks, Ubiquitous Computing, and Dependable Applications}, vol.~4, no.~1,
  pp. 98--117, 2013.

\bibitem{yerima2018droidfusion}
S.~Y. Yerima and S.~Sezer, ``Droidfusion: A novel multilevel classifier fusion
  approach for android malware detection,'' \emph{IEEE transactions on
  cybernetics}, vol.~49, no.~2, pp. 453--466, 2018.

\bibitem{grosse2016}
K.~Grosse, N.~Papernot, P.~Manoharan, M.~Backes, and P.~McDaniel,
  ``{Adversarial perturbations against deep neural networks for malware
  classification}.'' \emph{arXiv preprint arXiv:1606.04435}, 2016.

\bibitem{al2018adversarial}
A.~Al-Dujaili, A.~Huang, E.~Hemberg, and U.-M. O’Reilly, ``Adversarial deep
  learning for robust detection of binary encoded malware,'' in \emph{2018 IEEE
  Security and Privacy Workshops (SPW)}.\hskip 1em plus 0.5em minus 0.4em\relax
  IEEE, 2018, pp. 76--82.

\bibitem{rosenberg2018generic}
I.~Rosenberg, A.~Shabtai, L.~Rokach, and Y.~Elovici, ``Generic black-box
  end-to-end attack against state of the art api call based malware
  classifiers,'' in \emph{International Symposium on Research in Attacks,
  Intrusions, and Defenses}.\hskip 1em plus 0.5em minus 0.4em\relax Springer,
  2018, pp. 490--510.

\bibitem{demontis2017yes}
A.~Demontis, M.~Melis, B.~Biggio, D.~Maiorca, D.~Arp, K.~Rieck, I.~Corona,
  G.~Giacinto, and F.~Roli, ``Yes, machine learning can be more secure! a case
  study on android malware detection,'' \emph{IEEE Transactions on Dependable
  and Secure Computing}, 2017.

\bibitem{papernot2018sok}
N.~Papernot, P.~McDaniel, A.~Sinha, and M.~P. Wellman, ``Sok: Security and
  privacy in machine learning,'' in \emph{2018 IEEE European Symposium on
  Security and Privacy (EuroS\&P)}.\hskip 1em plus 0.5em minus 0.4em\relax
  IEEE, 2018, pp. 399--414.

\bibitem{goodfellow2014explaining}
I.~J. Goodfellow, J.~Shlens, and C.~Szegedy, ``Explaining and harnessing
  adversarial examples,'' \emph{arXiv preprint arXiv:1412.6572}, 2014.

\bibitem{grosse2017adversarial}
K.~Grosse, N.~Papernot, P.~Manoharan, M.~Backes, and P.~McDaniel, ``Adversarial
  examples for malware detection,'' in \emph{European symposium on research in
  computer security}.\hskip 1em plus 0.5em minus 0.4em\relax Springer, 2017,
  pp. 62--79.

\bibitem{yang2017malware}
W.~Yang, D.~Kong, T.~Xie, and C.~A. Gunter, ``Malware detection in adversarial
  settings: Exploiting feature evolutions and confusions in android apps,'' in
  \emph{Proceedings of the 33rd Annual Computer Security Applications
  Conference}, 2017, pp. 288--302.

\bibitem{kurakin2016adversarial}
A.~Kurakin, I.~Goodfellow, S.~Bengio \emph{et~al.}, ``Adversarial examples in
  the physical world,'' 2016.

\bibitem{carlini2017towards}
N.~Carlini and D.~Wagner, ``Towards evaluating the robustness of neural
  networks,'' in \emph{2017 ieee symposium on security and privacy (sp)}.\hskip
  1em plus 0.5em minus 0.4em\relax IEEE, 2017, pp. 39--57.

\bibitem{chen2020hopskipjumpattack}
J.~Chen, M.~I. Jordan, and M.~J. Wainwright, ``Hopskipjumpattack: A
  query-efficient decision-based attack,'' in \emph{2020 IEEE Symposium on
  Security and Privacy (SP)}.\hskip 1em plus 0.5em minus 0.4em\relax IEEE,
  2020, pp. 1277--1294.

\bibitem{madry2017towards}
A.~Madry, A.~Makelov, L.~Schmidt, D.~Tsipras, and A.~Vladu, ``Towards deep
  learning models resistant to adversarial attacks,'' \emph{arXiv preprint
  arXiv:1706.06083}, 2017.

\bibitem{papernot2016limitations}
N.~Papernot, P.~McDaniel, S.~Jha, M.~Fredrikson, Z.~B. Celik, and A.~Swami,
  ``The limitations of deep learning in adversarial settings,'' in \emph{2016
  IEEE European symposium on security and privacy (EuroS\&P)}.\hskip 1em plus
  0.5em minus 0.4em\relax IEEE, 2016, pp. 372--387.

\bibitem{10.1145/3473039}
\BIBentryALTinterwordspacing
L.~Demetrio, S.~E. Coull, B.~Biggio, G.~Lagorio, A.~Armando, and F.~Roli,
  ``Adversarial exemples: A survey and experimental evaluation of practical
  attacks on machine learning for windows malware detection,'' \emph{ACM Trans.
  Priv. Secur.}, vol.~24, no.~4, sep 2021. [Online]. Available:
  \url{https://doi.org/10.1145/3473039}
\BIBentrySTDinterwordspacing

\bibitem{rosenberg2020query}
I.~Rosenberg, A.~Shabtai, Y.~Elovici, and L.~Rokach, ``Query-efficient
  black-box attack against sequence-based malware classifiers,'' in
  \emph{Annual Computer Security Applications Conference}, 2020, pp. 611--626.

\bibitem{demetrio2019explaining}
L.~Demetrio, B.~Biggio, G.~Lagorio, F.~Roli, and A.~Armando, ``Explaining
  vulnerabilities of deep learning to adversarial malware binaries,''
  \emph{arXiv preprint arXiv:1901.03583}, 2019.

\bibitem{grosse2017statistical}
K.~Grosse, P.~Manoharan, N.~Papernot, M.~Backes, and P.~McDaniel, ``On the
  (statistical) detection of adversarial examples,'' \emph{arXiv preprint
  arXiv:1702.06280}, 2017.

\bibitem{ebrahimi2020binary}
M.~Ebrahimi, N.~Zhang, J.~Hu, M.~T. Raza, and H.~Chen, ``Binary black-box
  evasion attacks against deep learning-based static malware detectors with
  adversarial byte-level language model,'' \emph{arXiv preprint
  arXiv:2012.07994}, 2020.

\bibitem{sewak2021adversarialuscator}
M.~Sewak, S.~K. Sahay, and H.~Rathore, ``Adversarialuscator: An adversarial-drl
  based obfuscator and metamorphic malware swarm generator,'' in \emph{2021
  International Joint Conference on Neural Networks (IJCNN)}.\hskip 1em plus
  0.5em minus 0.4em\relax IEEE, 2021, pp. 1--9.

\bibitem{song2020mab}
\BIBentryALTinterwordspacing
W.~Song, X.~Li, S.~Afroz, D.~Garg, D.~Kuznetsov, and H.~Yin, ``Mab-malware: A
  reinforcement learning framework for blackbox generation of adversarial
  malware,'' in \emph{Proceedings of the 2022 ACM on Asia Conference on
  Computer and Communications Security}, ser. ASIA CCS '22.\hskip 1em plus
  0.5em minus 0.4em\relax New York, NY, USA: Association for Computing
  Machinery, 2022, p. 990–1003. [Online]. Available:
  \url{https://doi.org/10.1145/3488932.3497768}
\BIBentrySTDinterwordspacing

\bibitem{demetrio2021functionality}
L.~Demetrio, B.~Biggio, G.~Lagorio, F.~Roli, and A.~Armando,
  ``Functionality-preserving black-box optimization of adversarial windows
  malware,'' \emph{IEEE Transactions on Information Forensics and Security},
  vol.~16, pp. 3469--3478, 2021.

\bibitem{labaca2021universal}
R.~Labaca-Castro, L.~Mu{\~n}oz-Gonz{\'a}lez, F.~Pendlebury, G.~D. Rodosek,
  F.~Pierazzi, and L.~Cavallaro, ``Universal adversarial perturbations for
  malware,'' \emph{arXiv preprint arXiv:2102.06747}, 2021.

\bibitem{hu2017generating}
W.~Hu and Y.~Tan, ``Generating adversarial malware examples for black-box
  attacks based on gan,'' \emph{arXiv preprint arXiv:1702.05983}, 2017.

\bibitem{suciu2019exploring}
O.~Suciu, S.~E. Coull, and J.~Johns, ``Exploring adversarial examples in
  malware detection,'' in \emph{2019 IEEE Security and Privacy Workshops
  (SPW)}.\hskip 1em plus 0.5em minus 0.4em\relax IEEE, 2019, pp. 8--14.

\bibitem{demetrio2022practical}
L.~Demetrio, B.~Biggio, and F.~Roli, ``Practical attacks on machine learning: A
  case study on adversarial windows malware,'' \emph{IEEE Security \& Privacy},
  vol.~20, no.~5, pp. 77--85, 2022.

\bibitem{chen2019android}
X.~Chen, C.~Li, D.~Wang, S.~Wen, J.~Zhang, S.~Nepal, Y.~Xiang, and K.~Ren,
  ``Android hiv: A study of repackaging malware for evading machine-learning
  detection,'' \emph{IEEE Transactions on Information Forensics and Security},
  vol.~15, pp. 987--1001, 2019.

\bibitem{anderson2018learning}
H.~S. Anderson, A.~Kharkar, B.~Filar, D.~Evans, and P.~Roth, ``Learning to
  evade static pe machine learning malware models via reinforcement learning,''
  \emph{arXiv preprint arXiv:1801.08917}, 2018.

\bibitem{li2020enhancing}
D.~Li, Q.~Li, Y.~Ye, and S.~Xu, ``Enhancing deep neural networks against
  adversarial malware examples,'' \emph{arXiv preprint arXiv:2004.07919}, 2020.

\bibitem{chen2020training2}
Y.~Chen, S.~Wang, D.~She, and S.~Jana, ``On training robust $\{$PDF$\}$ malware
  classifiers,'' in \emph{29th USENIX Security Symposium (USENIX Security 20)},
  2020, pp. 2343--2360.

\bibitem{maiorca2019towards}
D.~Maiorca, B.~Biggio, and G.~Giacinto, ``Towards adversarial malware
  detection: Lessons learned from pdf-based attacks,'' \emph{ACM Computing
  Surveys (CSUR)}, vol.~52, no.~4, pp. 1--36, 2019.

\bibitem{10.1145/3484491}
\BIBentryALTinterwordspacing
D.~Li, Q.~Li, Y.~F. Ye, and S.~Xu, ``Arms race in adversarial malware
  detection: A survey,'' \emph{ACM Comput. Surv.}, vol.~55, no.~1, nov 2021.
  [Online]. Available: \url{https://doi.org/10.1145/3484491}
\BIBentrySTDinterwordspacing

\bibitem{ling2023adversarial}
X.~Ling, L.~Wu, J.~Zhang, Z.~Qu, W.~Deng, X.~Chen, Y.~Qian, C.~Wu, S.~Ji,
  T.~Luo \emph{et~al.}, ``Adversarial attacks against windows pe malware
  detection: A survey of the state-of-the-art,'' \emph{Computers \& Security},
  p. 103134, 2023.

\bibitem{aryal2021survey}
K.~Aryal, M.~Gupta, and M.~Abdelsalam, ``A survey on adversarial attacks for
  malware analysis,'' \emph{arXiv preprint arXiv:2111.08223}, 2021.

\bibitem{li2021arms}
D.~Li, Q.~Li, Y.~Ye, and S.~Xu, ``Arms race in adversarial malware detection: A
  survey,'' \emph{ACM Computing Surveys (CSUR)}, vol.~55, no.~1, pp. 1--35,
  2021.

\bibitem{carlini2019evaluating}
N.~Carlini, A.~Athalye, N.~Papernot, W.~Brendel, J.~Rauber, D.~Tsipras,
  I.~Goodfellow, A.~Madry, and A.~Kurakin, ``On evaluating adversarial
  robustness,'' \emph{arXiv preprint arXiv:1902.06705}, 2019.

\bibitem{stokes2017attack}
J.~W. Stokes, D.~Wang, M.~Marinescu, M.~Marino, and B.~Bussone, ``Attack and
  defense of dynamic analysis-based, adversarial neural malware classification
  models,'' \emph{arXiv preprint arXiv:1712.05919}, 2017.

\bibitem{athalye2018obfuscated}
A.~Athalye, N.~Carlini, and D.~Wagner, ``Obfuscated gradients give a false
  sense of security: Circumventing defenses to adversarial examples,'' in
  \emph{International conference on machine learning}.\hskip 1em plus 0.5em
  minus 0.4em\relax PMLR, 2018, pp. 274--283.

\bibitem{papernot2017practical}
N.~Papernot, P.~McDaniel, I.~Goodfellow, S.~Jha, Z.~Celik, and A.~Swami,
  ``{Practical black-box attacks against machine learning}.'' \emph{Proceedings
  of the 2017 ACM on Asia conference on computer and communications security},
  pp. 506--519, 2018.

\bibitem{severi2021explanation}
G.~Severi, J.~Meyer, S.~Coull, and A.~Oprea, ``Explanation-guided backdoor
  poisoning attacks against malware classifiers,'' in \emph{30th $\{$USENIX$\}$
  Security Symposium ($\{$USENIX$\}$ Security 21)}, 2021.

\bibitem{biggio2013evasion}
B.~Biggio, I.~Corona, D.~Maiorca, B.~Nelson, N.~{\v{S}}rndi{\'c}, P.~Laskov,
  G.~Giacinto, and F.~Roli, ``Evasion attacks against machine learning at test
  time,'' in \emph{Joint European conference on machine learning and knowledge
  discovery in databases}.\hskip 1em plus 0.5em minus 0.4em\relax Springer,
  2013, pp. 387--402.

\bibitem{apruzzese2023realgradients}
G.~Apruzzese, H.~S. Anderson, S.~Dambra, D.~Freeman, F.~Pierazzi, and K.~A.
  Roundy, ``"real attackers don't compute gradients": Bridging the gap between
  adversarial ml research and practice,'' in \emph{Proceedings of the 1st IEEE
  Conference on Secure and Trustworthy Machine Learning (SaTML)}, 2023.

\bibitem{laskov2014practical}
P.~Laskov \emph{et~al.}, ``Practical evasion of a learning-based classifier: A
  case study,'' in \emph{2014 IEEE symposium on security and privacy}.\hskip
  1em plus 0.5em minus 0.4em\relax IEEE, 2014, pp. 197--211.

\bibitem{santana2021detecting}
E.~J. Santana, R.~P. Silva, B.~B. Zarpel{\~a}o, and S.~Barbon~Junior,
  ``Detecting and mitigating adversarial examples in regression tasks: A
  photovoltaic power generation forecasting case study,'' \emph{Information},
  vol.~12, no.~10, p. 394, 2021.

\bibitem{biggio2018wild}
B.~Biggio and F.~Roli, ``Wild patterns: Ten years after the rise of adversarial
  machine learning,'' \emph{Pattern Recognition}, vol.~84, pp. 317--331, 2018.

\bibitem{carlini2017adversarial}
N.~Carlini and D.~Wagner, ``Adversarial examples are not easily detected:
  Bypassing ten detection methods,'' in \emph{Proceedings of the 10th ACM
  workshop on artificial intelligence and security}, 2017, pp. 3--14.

\bibitem{217486}
O.~Suciu, R.~Marginean, Y.~Kaya, H.~D. III, and T.~Dumitras, ``When does
  machine learning {FAIL}? generalized transferability for evasion and
  poisoning attacks,'' in \emph{27th {USENIX} Security Symposium ({USENIX}
  Security 18)}.\hskip 1em plus 0.5em minus 0.4em\relax Baltimore, MD: {USENIX}
  Association, Aug. 2018, pp. 1299--1316.

\bibitem{shu2018threat}
X.~Shu, F.~Araujo, D.~L. Schales, M.~P. Stoecklin, J.~Jang, H.~Huang, and J.~R.
  Rao, ``Threat intelligence computing,'' in \emph{Proceedings of the 2018 ACM
  SIGSAC Conference on Computer and Communications Security}, 2018, pp.
  1883--1898.

\bibitem{zhu2018chainsmith}
Z.~Zhu and T.~Dumitras, ``Chainsmith: Automatically learning the semantics of
  malicious campaigns by mining threat intelligence reports,'' in \emph{2018
  IEEE European Symposium on Security and Privacy (EuroS\&P)}.\hskip 1em plus
  0.5em minus 0.4em\relax IEEE, 2018, pp. 458--472.

\bibitem{ilyas2018black}
A.~Ilyas, L.~Engstrom, A.~Athalye, and J.~Lin, ``Black-box adversarial attacks
  with limited queries and information,'' in \emph{International Conference on
  Machine Learning}.\hskip 1em plus 0.5em minus 0.4em\relax PMLR, 2018, pp.
  2137--2146.

\bibitem{papernot2016transferability}
N.~Papernot, P.~McDaniel, and I.~Goodfellow, ``Transferability in machine
  learning: from phenomena to black-box attacks using adversarial samples,''
  \emph{arXiv preprint arXiv:1605.07277}, 2016.

\bibitem{brendel2017decision}
W.~Brendel, J.~Rauber, and M.~Bethge, ``Decision-based adversarial attacks:
  Reliable attacks against black-box machine learning models,'' \emph{arXiv
  preprint arXiv:1712.04248}, 2017.

\bibitem{chen2020stateful}
S.~Chen, N.~Carlini, and D.~Wagner, ``Stateful detection of black-box
  adversarial attacks,'' in \emph{Proceedings of the 1st ACM Workshop on
  Security and Privacy on Artificial Intelligence}, 2020, pp. 30--39.

\bibitem{Li_2020_CVPR}
H.~Li, X.~Xu, X.~Zhang, S.~Yang, and B.~Li, ``Qeba: Query-efficient
  boundary-based blackbox attack,'' in \emph{CVPR}, 2020.

\bibitem{paruchuri2008playing}
P.~Paruchuri, J.~P. Pearce, J.~Marecki, M.~Tambe, F.~Ordonez, and S.~Kraus,
  ``Playing games for security: An efficient exact algorithm for solving
  bayesian stackelberg games,'' in \emph{Proceedings of the 7th international
  joint conference on Autonomous agents and multiagent systems-Volume 2}.\hskip
  1em plus 0.5em minus 0.4em\relax International Foundation for Autonomous
  Agents and Multiagent Systems, 2008, pp. 895--902.

\bibitem{tambe2011security}
M.~Tambe, \emph{Security and game theory: algorithms, deployed systems, lessons
  learned}.\hskip 1em plus 0.5em minus 0.4em\relax Cambridge university press,
  2011.

\bibitem{gurobi}
\BIBentryALTinterwordspacing
{Gurobi Optimization, LLC}, ``{Gurobi Optimizer Reference Manual},'' 2022.
  [Online]. Available: \url{https://www.gurobi.com}
\BIBentrySTDinterwordspacing

\bibitem{arp2022and}
D.~Arp, E.~Quiring, F.~Pendlebury, A.~Warnecke, F.~Pierazzi, C.~Wressnegger,
  L.~Cavallaro, and K.~Rieck, ``Dos and don’ts of machine learning in
  computer security,'' in \emph{Proc. of the USENIX Security Symposium}, 2022.

\bibitem{8949524}
O.~A. Aslan and R.~Samet, ``A comprehensive review on malware detection
  approaches,'' \emph{IEEE Access}, vol.~8, pp. 6249--6271, 2020.

\bibitem{arp2014drebin}
D.~Arp, M.~Spreitzenbarth, M.~Hubner, H.~Gascon, K.~Rieck, and C.~Siemens,
  ``Drebin: Effective and explainable detection of android malware in your
  pocket.'' in \emph{Ndss}, vol.~14, 2014, pp. 23--26.

\bibitem{LIEF}
R.~Thomas, ``Lief - library to instrument executable formats,''
  https://lief.quarkslab.com/, April 2017.

\bibitem{ma2021partner}
J.~Ma, H.~Xie, G.~Han, S.-F. Chang, A.~Galstyan, and W.~Abd-Almageed,
  ``Partner-assisted learning for few-shot image classification,'' in
  \emph{Proceedings of the IEEE/CVF International Conference on Computer
  Vision}, 2021, pp. 10\,573--10\,582.

\bibitem{ravi2016optimization}
S.~Ravi and H.~Larochelle, ``Optimization as a model for few-shot learning,''
  2016.

\bibitem{li2019few}
A.~Li, T.~Luo, T.~Xiang, W.~Huang, and L.~Wang, ``Few-shot learning with global
  class representations,'' in \emph{Proceedings of the IEEE/CVF International
  Conference on Computer Vision}, 2019, pp. 9715--9724.

\bibitem{yao2019hierarchically}
H.~Yao, Y.~Wei, J.~Huang, and Z.~Li, ``Hierarchically structured
  meta-learning,'' in \emph{International Conference on Machine
  Learning}.\hskip 1em plus 0.5em minus 0.4em\relax PMLR, 2019, pp. 7045--7054.

\bibitem{du2021metakernel}
Y.~Du, H.~Sun, X.~Zhen, J.~Xu, Y.~Yin, L.~Shao, and C.~G. Snoek, ``Metakernel:
  Learning variational random features with limited labels,'' \emph{arXiv
  preprint arXiv:2105.03781}, 2021.

\bibitem{ye2021train}
H.-J. Ye and W.-L. Chao, ``How to train your maml to excel in few-shot
  classification,'' \emph{arXiv preprint arXiv:2106.16245}, 2021.

\bibitem{6234405}
C.~Rossow, C.~J. Dietrich, C.~Grier, C.~Kreibich, V.~Paxson, N.~Pohlmann,
  H.~Bos, and M.~v. Steen, ``Prudent practices for designing malware
  experiments: Status quo and outlook,'' in \emph{2012 IEEE Symposium on
  Security and Privacy}, 2012, pp. 65--79.

\bibitem{scikit-learn}
F.~Pedregosa, G.~Varoquaux, A.~Gramfort, V.~Michel, B.~Thirion, O.~Grisel,
  M.~Blondel, P.~Prettenhofer, R.~Weiss, V.~Dubourg, J.~Vanderplas, A.~Passos,
  D.~Cournapeau, M.~Brucher, M.~Perrot, and E.~Duchesnay, ``Scikit-learn:
  Machine learning in {P}ython,'' \emph{Journal of Machine Learning Research},
  vol.~12, pp. 2825--2830, 2011.

\bibitem{chollet2015keras}
F.~Chollet \emph{et~al.}, ``Keras,'' \url{https://keras.io}, 2015.

\bibitem{tensorflow2015-whitepaper}
\BIBentryALTinterwordspacing
{Abadi et al.}, ``{TensorFlow}: Large-scale machine learning on heterogeneous
  systems,'' 2015. [Online]. Available: \url{https://www.tensorflow.org/}
\BIBentrySTDinterwordspacing

\bibitem{jackson2010multi}
T.~Jackson, C.~Wimmer, and M.~Franz, ``Multi-variant program execution for
  vulnerability detection and analysis,'' in \emph{Proceedings of the Sixth
  Annual Workshop on Cyber Security and Information Intelligence Research},
  2010, pp. 1--4.

\bibitem{jackson2010effectiveness}
T.~Jackson, B.~Salamat, G.~Wagner, C.~Wimmer, and M.~Franz, ``On the
  effectiveness of multi-variant program execution for vulnerability detection
  and prevention,'' in \emph{Proceedings of the 6th International Workshop on
  Security Measurements and Metrics}, 2010, pp. 1--8.

\bibitem{moser2007limits}
A.~Moser, C.~Kruegel, and E.~Kirda, ``Limits of static analysis for malware
  detection,'' in \emph{Twenty-third annual computer security applications
  conference (ACSAC 2007)}.\hskip 1em plus 0.5em minus 0.4em\relax IEEE, 2007,
  pp. 421--430.

\bibitem{li2021framework}
D.~Li, Q.~Li, Y.~Ye, and S.~Xu, ``A framework for enhancing deep neural
  networks against adversarial malware,'' \emph{IEEE Transactions on Network
  Science and Engineering}, vol.~8, no.~1, pp. 736--750, 2021.

\bibitem{8171381}
Z.~Abaid, M.~A. Kaafar, and S.~Jha, ``Quantifying the impact of adversarial
  evasion attacks on machine learning based android malware classifiers,'' in
  \emph{2017 IEEE 16th International Symposium on Network Computing and
  Applications (NCA)}, 2017, pp. 1--10.

\bibitem{moosavi2016deepfool}
S.-M. Moosavi-Dezfooli, A.~Fawzi, and P.~Frossard, ``Deepfool: a simple and
  accurate method to fool deep neural networks,'' in \emph{Proceedings of the
  IEEE conference on computer vision and pattern recognition}, 2016, pp.
  2574--2582.

\bibitem{dhaliwal2018gradient}
J.~Dhaliwal and S.~Shintre, ``Gradient similarity: An explainable approach to
  detect adversarial attacks against deep learning,'' \emph{arXiv preprint
  arXiv:1806.10707}, 2018.

\bibitem{moosavi2017universal}
S.-M. Moosavi-Dezfooli, A.~Fawzi, O.~Fawzi, and P.~Frossard, ``Universal
  adversarial perturbations,'' in \emph{Proceedings of the IEEE conference on
  computer vision and pattern recognition}, 2017, pp. 1765--1773.

\bibitem{bucilua2006model}
C.~Buciluǎ, R.~Caruana, and A.~Niculescu-Mizil, ``Model compression,'' in
  \emph{Proceedings of the 12th ACM SIGKDD international conference on
  Knowledge discovery and data mining}, 2006, pp. 535--541.

\bibitem{8594826}
S.~Buschjager, K.-H. Chen, J.-J. Chen, and K.~Morik, ``Realization of random
  forest for real-time evaluation through tree framing,'' in \emph{2018 IEEE
  International Conference on Data Mining (ICDM)}, 2018, pp. 19--28.

\bibitem{tang2018random}
C.~Tang, D.~Garreau, and U.~von Luxburg, ``When do random forests fail?'' in
  \emph{NeurIPS}, 2018, pp. 2987--2997.

\end{thebibliography}

\begin{appendices}

\section{Architecture of vanilla models}
\label{appendix:architecutres}
The following vanilla models are used in some instances (Section~\ref{sec:experimentalsetup}).
\begin{table}[H]
\centering
\scalebox{0.75}{
\begin{tabular}{ll} 
\hline
Model & Parameters \\
\hline
Decision Tree &  max\_depth=5, min\_samples\_leaf=1 \\ 
\hdashline
Neural Network & \makecell[l]{4 fully-connected layers (128 (Relu), \\  64 (Relu), 32 (Relu), 2 (Softmax))} \\
\hdashline
Random Forest &  max\_depth=100 \\ 
\hdashline
Support Vector Machine &  LinearSVC with probability enabled \\ 
\hline
\end{tabular}
}
\caption{Architectures of vanilla models.}
\end{table}

\section{StratDef strategies}
\label{appendix:mtdstrategies}
5 attacker types, 2 datasets and 6 StratDef configurations, leads to 60 strategy vectors. For brevity, we only include some examples of the StratDef strategies for both datasets. For each strategy vector, the rows correspond to the models selected through our model selection methods (Best \& Variety). Within each row, the probability of that model being selected at a particular attack intensity is listed.

\begin{table}[!htbp]
\centering
\scalebox{0.45}{
\begin{tabular}{llllllllllll}
\hline
 & 0 & 0.1 & 0.2 & 0.3 & 0.4 & 0.5 & 0.6 & 0.7 & 0.8 & 0.9 & 1 \\
\hline
DT-AT-0.01 & 0 & 0 & 0 & 0 & 0 & 0 & 0 & 0 & 0 & 0 & 0 \\
DT-AT-0.1 & 0 & 0 & 0 & 0 & 0 & 0.985507 & 0.985507 & 0.985507 & 0.985507 & 0.985507 & 0.985507 \\
DT-RFN & 0 & 0 & 0 & 0.510791 & 0.510791 & 0.014493 & 0.014493 & 0.014493 & 0.014493 & 0.014493 & 0.014493 \\
NN-RFN & 0 & 0 & 0 & 0 & 0 & 0 & 0 & 0 & 0 & 0 & 0 \\
RF & 0 & 0 & 0 & 0 & 0 & 0 & 0 & 0 & 0 & 0 & 0 \\
RF-AT-0.25 & 1 & 1 & 1 & 0.489209 & 0.489209 & 0 & 0 & 0 & 0 & 0 & 0 \\
RF-RFN & 0 & 0 & 0 & 0 & 0 & 0 & 0 & 0 & 0 & 0 & 0 \\
SECSVM-AT-0.25 & 0 & 0 & 0 & 0 & 0 & 0 & 0 & 0 & 0 & 0 & 0 \\
SVM & 0 & 0 & 0 & 0 & 0 & 0 & 0 & 0 & 0 & 0 & 0 \\
\hline
\end{tabular}
}

\caption{DREBIN, StratDef-Best-GT, Strong attacker}
\end{table}

\begin{table}[!htbp]
\centering
\scalebox{0.4}{
\begin{tabular}{llllllllllll}
\hline
 & 0 & 0.1 & 0.2 & 0.3 & 0.4 & 0.5 & 0.6 & 0.7 & 0.8 & 0.9 & 1 \\
\hline
DT-AT-0.01 & 0 & 0 & 0 & 0 & 0 & 0 & 0 & 0 & 0 & 0 & 0 \\
DT-AT-0.1 & 0 & 0 & 0.133333 & 0.133333 & 0.266667 & 0.266667 & 0.266667 & 0.266667 & 0.333333 & 0.333333 & 0.333333 \\
DT-RFN & 0 & 0 & 0 & 0 & 0.066667 & 0.066667 & 0.066667 & 0.066667 & 0.066667 & 0.066667 & 0.2 \\
NN-RFN & 0 & 0.066667 & 0 & 0 & 0 & 0 & 0 & 0 & 0 & 0 & 0 \\
RF & 0 & 0.266667 & 0.2 & 0.266667 & 0.2 & 0.2 & 0.2 & 0.133333 & 0.2 & 0.133333 & 0.066667 \\
RF-AT-0.25 & 1 & 0.333333 & 0.333333 & 0.333333 & 0.333333 & 0.333333 & 0.333333 & 0.333333 & 0.266667 & 0.266667 & 0.266667 \\
RF-RFN & 0 & 0.2 & 0.266667 & 0.2 & 0.133333 & 0.133333 & 0.133333 & 0.2 & 0.133333 & 0.2 & 0.133333 \\
SECSVM-AT-0.25 & 0 & 0.133333 & 0.066667 & 0.066667 & 0 & 0 & 0 & 0 & 0 & 0 & 0 \\
SVM & 0 & 0 & 0 & 0 & 0 & 0 & 0 & 0 & 0 & 0 & 0 \\
\hline
\end{tabular}
}

\caption{DREBIN, StratDef-Best-Ranked, Strong attacker}
\end{table}

\begin{table}[!htbp]
\centering
\scalebox{0.5}{
\begin{tabular}{llllllllllll}
\hline
 & 0 & 0.1 & 0.2 & 0.3 & 0.4 & 0.5 & 0.6 & 0.7 & 0.8 & 0.9 & 1 \\
\hline
DT\_AT\_0.01 & 0 & 0 & 0 & 0 & 0 & 0 & 0 & 0 & 0 & 0 & 0 \\
DT\_AT\_0.1 & 0 & 0 & 0 & 0 & 0 & 0 & 0.890756 & 0.890756 & 0.890756 & 0.890756 & 0.897638 \\
DT\_AT\_0.25 & 0 & 0 & 0 & 0 & 0 & 0 & 0 & 0 & 0 & 0 & 0 \\
NN & 0 & 0 & 0 & 0 & 0 & 0 & 0 & 0 & 0 & 0 & 0 \\
NN\_AT\_0.25 & 0 & 0 & 0 & 0 & 0 & 0 & 0 & 0 & 0 & 0 & 0.102362 \\
RF\_AT\_0.25 & 1 & 1 & 1 & 1 & 1 & 1 & 0.109244 & 0.109244 & 0.109244 & 0.109244 & 0 \\
SECSVM\_AT\_0.05 & 0 & 0 & 0 & 0 & 0 & 0 & 0 & 0 & 0 & 0 & 0 \\
SECSVM\_AT\_0.25 & 0 & 0 & 0 & 0 & 0 & 0 & 0 & 0 & 0 & 0 & 0 \\
SVM & 0 & 0 & 0 & 0 & 0 & 0 & 0 & 0 & 0 & 0 & 0 \\
\hline
\end{tabular}
}

\caption{DREBIN, StratDef-Variety-GT, Weak attacker}
\end{table}

\begin{table}[!htbp]
\centering
\scalebox{0.6}{
\begin{tabular}{llllllllllll}
\hline
 & 0 & 0.1 & 0.2 & 0.3 & 0.4 & 0.5 & 0.6 & 0.7 & 0.8 & 0.9 & 1 \\
\hline
DT\_AT\_0.01 & 0 & 0 & 0 & 0 & 0 & 0 & 0 & 0 & 0 & 0 & 0 \\
DT\_AT\_0.1 & 0 & 0.1 & 0.2 & 0.3 & 0.3 & 0.3 & 0.3 & 0.3 & 0.3 & 0.3 & 0.3 \\
NN\_AT\_0.25 & 0 & 0 & 0.1 & 0.1 & 0.1 & 0.1 & 0.1 & 0.1 & 0.1 & 0.1 & 0.1 \\
NN\_RFN & 0 & 0.2 & 0 & 0 & 0 & 0 & 0 & 0 & 0 & 0 & 0 \\
RF\_AT\_0.25 & 1 & 0.4 & 0.4 & 0.4 & 0.4 & 0.4 & 0.4 & 0.4 & 0.4 & 0.4 & 0.4 \\
SECSVM\_AT\_0.25 & 0 & 0.3 & 0.3 & 0.2 & 0.2 & 0.2 & 0.2 & 0.2 & 0.2 & 0.2 & 0.2 \\
SVM & 0 & 0 & 0 & 0 & 0 & 0 & 0 & 0 & 0 & 0 & 0 \\
\hline
\end{tabular}
}

\caption{DREBIN, StratDef-Variety-Ranked, Random attacker}
\end{table}

\begin{table}[!htbp]
\centering
\scalebox{0.5}{
\begin{tabular}{llllllllllll}
\hline
 & 0 & 0.1 & 0.2 & 0.3 & 0.4 & 0.5 & 0.6 & 0.7 & 0.8 & 0.9 & 1 \\
\hline
DT\_AT\_0.1 & 0 & 0 & 0 & 0 & 0 & 0 & 0 & 0 & 0 & 0 & 0 \\
DT\_AT\_0.25 & 0 & 0 & 0 & 0 & 0 & 0 & 0 & 0 & 0 & 0 & 0 \\
NN & 1 & 0 & 0 & 0 & 0 & 0 & 0 & 0 & 0 & 0 & 0 \\
NN\_AT\_0.01 & 0 & 0 & 0 & 0 & 0 & 0 & 0 & 0 & 0 & 0 & 0 \\
NN\_AT\_0.1 & 0 & 0 & 0 & 0 & 0 & 0 & 0 & 0 & 0 & 0 & 0 \\
NN\_AT\_0.25 & 0 & 0 & 0 & 0 & 0 & 0 & 0 & 0 & 0 & 0 & 0.333333 \\
RF & 0 & 0 & 0 & 0 & 0 & 0 & 0 & 0 & 0 & 0 & 0 \\
RF\_AT\_0.1 & 0 & 0.5 & 0.5 & 0.5 & 0.5 & 0.5 & 0.5 & 0.5 & 0.5 & 0.5 & 0.333333 \\
RF\_AT\_0.25 & 0 & 0.5 & 0.5 & 0.5 & 0.5 & 0.5 & 0.5 & 0.5 & 0.5 & 0.5 & 0.333333 \\
RF\_RFN & 0 & 0 & 0 & 0 & 0 & 0 & 0 & 0 & 0 & 0 & 0 \\
SECSVM\_AT\_0.1 & 0 & 0 & 0 & 0 & 0 & 0 & 0 & 0 & 0 & 0 & 0 \\
SECSVM\_AT\_0.25 & 0 & 0 & 0 & 0 & 0 & 0 & 0 & 0 & 0 & 0 & 0 \\
\hline
\end{tabular}
}

\caption{SLEIPNIR, StratDef-Best-GT, Medium attacker}

\end{table}

\section{FPR of majority voting vs. veto voting}
\label{appendix:votingfpr}
\begin{figure}[H]
\centering
\begin{tikzpicture}[scale=0.5]
    \begin{axis}[
        ybar,
        enlargelimits=0.15,
        ylabel={FPR (\%)},
        symbolic x coords={DREBIN, SLEIPNIR},
        xtick=data,
        nodes near coords align={horizontal},
        legend pos=outer north east,
        legend style={nodes={scale=0.5, transform shape}},
        height=4cm,
        width=15cm
        ]
    \addplot coordinates {(DREBIN, 4.1) (SLEIPNIR, 11.2) };
    \addplot coordinates {(DREBIN, 25.8) (SLEIPNIR, 21.7) };
    
    \legend{Voting-Variety-Majority, Voting-Variety-Veto}
    \end{axis}
   
\end{tikzpicture}
\caption{Average FPR of Voting-Variety-Majority \& Voting-Variety-Veto against the strong attacker across intensities.}
\label{figure:votingvsfpr}
\end{figure}
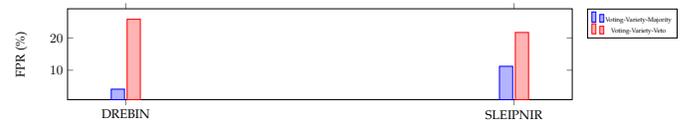

\section{Extended results}
\label{appendix:extendedresults}

The extended results are located in the following repository: \url{https://osf.io/93yzt/?view_only=bac46b0ab58b42758a133ac48f36b017}

Note that AUC and FPR require two classes. At \(\alpha = 1\), there is only one class (malware) and therefore the values of these metrics are undefined or ``nan'' at this attack intensity. If these metrics are used in the consideration score, for \(\alpha=1\), we use average value of these metrics across all other attack intensities instead.

\end{appendices}

\end{document}